\documentclass{article}

\usepackage[final]{neurips_2023}

\usepackage{amsmath,amsfonts,bm}

\def\eqref#1{equation~\ref{#1}}

\def\1{\bm{1}}

\DeclareMathAlphabet{\mathsfit}{\encodingdefault}{\sfdefault}{m}{sl}
\SetMathAlphabet{\mathsfit}{bold}{\encodingdefault}{\sfdefault}{bx}{n}

\usepackage[utf8]{inputenc} %
\usepackage[T1]{fontenc}    %
\usepackage{hyperref}       %
\usepackage{url}            %
\usepackage{booktabs}       %
\usepackage{amsfonts}       %
\usepackage{nicefrac}       %
\usepackage{microtype}      %
\usepackage{xcolor}         %
\usepackage{wrapfig}
\usepackage{wasysym}
\usepackage{marvosym}     %
\usepackage{paralist}
\usepackage{amsmath}
\usepackage{amssymb}
\usepackage{mathtools}
\usepackage{amsthm}

\usepackage{times}
\usepackage{latexsym}
\usepackage{cleveref}
\usepackage{graphicx}
\usepackage{subcaption}

\usepackage[colorinlistoftodos,prependcaption,textsize=tiny,disable]{todonotes}
\usepackage{algorithm}
\usepackage{algorithmic}
\usepackage[export]{adjustbox}
\usepackage{scalerel,xparse}
\renewcommand{\algorithmicrequire}{\textbf{Input:}}
\renewcommand{\algorithmicensure}{\textbf{Output:}}

\title{Improving Language Plasticity via \\ Pretraining with Active Forgetting}

\author{Yihong Chen$^{\text{\Aries \Taurus}}$ \quad Kelly Marchisio$^{\text{\Leo}}$ \quad Roberta Raileanu$^{\text{\Taurus}}$ \quad David Ifeoluwa Adelani$^{\text{\Aries}}$ \\ 
\bf Pontus Stenetorp$^{\text{\Aries}}$ \quad Sebastian Riedel$^{\text{\Aries}}$ \quad Mikel Artetxe$^{\text{\Aquarius}}$ \\
$^{\text{\Aries}}$UCL Centre for Artificial Intelligence \\
$^{\text{\Taurus}}$Meta AI \quad 
$^{\text{\Aquarius}}$Reka AI \qquad
$^{\text{\Leo}}$Cohere AI \\
\texttt{\{yihong.chen, d.adelani, p.stenetorp, s.riedel\}@cs.ucl.ac.uk} \\
\texttt{mikel@reka.ai} \quad \texttt{kelly@cohere.com} \quad \texttt{raileanu@meta.com} \\
}

\begin{document}

\maketitle

\begin{abstract}
Pretrained language models~(PLMs) are today the primary model for natural language processing. 
Despite their impressive downstream performance, it can be difficult to apply PLMs to new languages, a barrier to making their capabilities universally accessible.
While prior work has shown it possible to address this issue by learning a new embedding layer for the new language, doing so is both data and compute inefficient.
We propose to use an \textit{active forgetting} mechanism during pretraining, as a simple way of creating PLMs that can quickly adapt to new languages.
Concretely, by resetting the embedding layer every $K$ updates during pretraining, we encourage the PLM to improve its ability of learning new embeddings within limited number of updates, similar to a meta-learning effect.
Experiments with RoBERTa show that models pretrained with our forgetting mechanism not only demonstrate faster convergence during language adaptation, but also outperform standard ones in a low-data regime, particularly for languages that are distant from English. Code will be available at \url{https://github.com/facebookresearch/language-model-plasticity}.
\end{abstract}

\section{Introduction}
Pretrained language models~(PLMs) have been swiftly reshaping the landscape of natural language processing~(NLP) by improving upon standardized benchmarks across the board~\citep{Radford2018ImprovingLU,devlin-etal-2019-bert,roberta,GPT3}. At their core, they acquire knowledge by ingesting large datasets and store this knowledge in their parameters during pretraining. Using finetuning or prompting~\citep{GPT3}, such knowledge can then be applied to downstream applications, such as semantic analysis, question answering, and others.

Despite their success, PLMs still have a number of shortcomings~\citep{dm_ethical_social_risks,weidinger2022taxonomy}. In particular, it requires massive data and computation to pretrain them~\citep{gururangan2020don,kaplan2020scaling,hernandez2021scaling,hulora,touvron2023llama}.
Naively retraining a new PLM to accommodate every lingual space shift ~\footnote{We use the term \emph{lingual space shift} to describe changes in language usage between pretraining and the target downstream application, caused by factors such as language change, time evolution, or domain switch. A model with high \emph{language plasticity} would quickly adapt to these shifts.} would be prohibitively expensive. 
This makes it a highly relevant research target to create PLMs that can be efficiently adapted to new lingual spaces.

While forgetting in the context of both human and machine learning is often perceived as something negative~(for instance catastrophic forgetting~\citep{MCCLOSKEY1989109,ratcliff1990connectionist,kirkpatrick2017overcoming}), recent works have shown that for artificial neural networks forgetting can also play a positive role in increasing their ``plasticity'', such as improving generalization to unseen data~\citep{zhoufortuitous,chenrefactor,igl2021transient}, enabling learning in low-data regimes~\citep{alabdulmohsin2021impact,taha2021knowledge}, or counteracting primacy bias~\citep{nikishin2022primacy,d2022sample}. Given these developments, in this work we ask whether we can draw upon forgetting as a mechanism to improve \emph{pretraining} and imbue PLMs with similar benefits.

\begin{wrapfigure}{L}{0.618\textwidth}
  \begin{center}
  \includegraphics[width=0.618\textwidth]{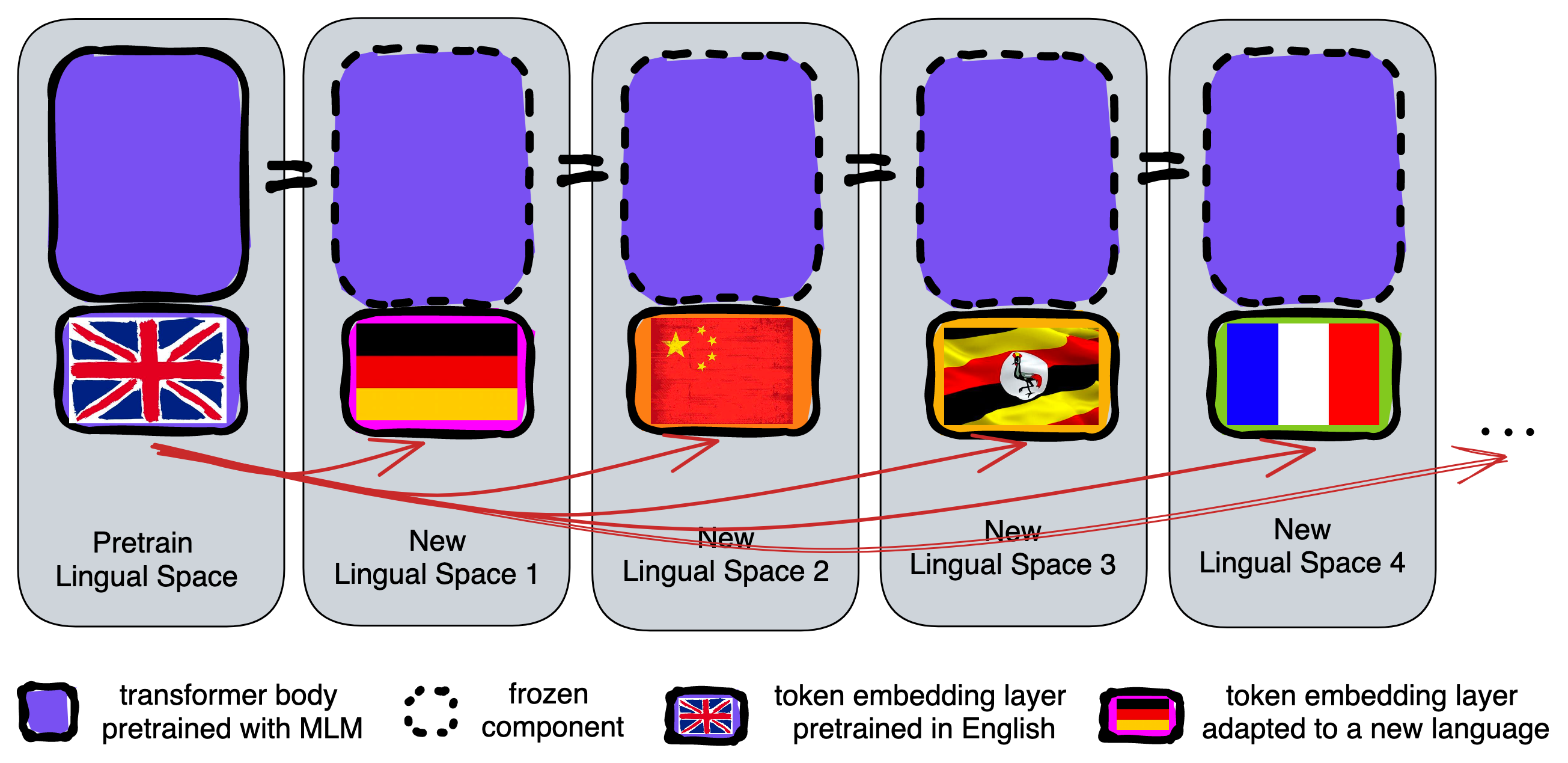}
  \end{center}
    \caption{
        \textit{Rewiring} via relearning token embeddings: where the transformer body (the purple part) is ``frozen'' and reused for a new language, but the token embeddings are relearned to suit the new language.
    }
    \label{fig:recharge}
\end{wrapfigure}

It is well established in the NLP community that models struggle to generalize across languages without substantial intervention~\citep{conneau-etal-2020-unsupervised,pfeiffer-etal-2020-mad,pfeiffer2022lifting,ansell2022composable}, which is especially true for low-resources languages. We thus see this as a promising testing ground for forgetting techniques. Our focus is on the input layer of the PLM, the \textit{token embedding layer}, as learning it has been shown to be highly effective when adapting between languages~\citep{artetxe2020cross}.

Concretely, we introduce an \textit{active forgetting} mechanism, that resets token embeddings at regular intervals, while leaving all other parameters untouched throughout pretraining. We study whether this forgetting approach creates a PLM that can easily \textit{rewire}~(Figure~\ref{fig:recharge}) to an unseen (possibly distant) language. Intuitively, resetting embeddings forces the transformer body to re-derive reasoning each time instead of relying on memorized shortcuts. Through repetition, the body learns more abstract, high-level reasoning. A model with greater abstraction can easily transfer across languages, since high-level reasoning is more language-agnostic.

Our zero-shot evaluations on several cross-lingual transfer benchmarks show that for cases where unlabeled adaptation corpus for the unseen language has as few as $5$ million tokens (a low-data regime), forgetting PLMs outperforms the baseline by large margins: average gains of {\color{blue}$+21.2\%$} on XNLI, {\color{blue}$+33.8\%$} on MLQA, and {\color{blue}$+60.9\%$} on XQuAD. In addition, models pretrained using active forgetting converge faster during language adaptation. Finally, we find that forgetting is especially beneficial for languages that \emph{are distant from} English, such as Arabic, Hindi, Thai, and Turkish.

\section{Rewire PLMs for New Languages}
\label{sec:bg}
Using unlabeled data, \citet{artetxe2020cross} demonstrates the possibility of rewiring a monolingual PLM for a new language; they propose to relearn the embedding layer for the new language while keeping all the other parameters frozen. The underlying assumption is that the token embedding layer and the transformer body~(the non-token-embedding parameters) divide up the responsibility in a way that the former handles language-specific lexical meanings, while the latter deals with high-level general reasoning. Hence, rewiring an English PLM for a new language boils down to separately adapting the former with unlabeled data in the new language and the latter with English task data. The procedure can be summarized as follows:
\begin{enumerate}
    \item{ {\color{violet}Pretrain}: A transformer-based model is pretrained on an \textit{English} corpus.
    In our experiments, we choose to pretrain RoBERTa-base~\cite{roberta}, a 12-layer transformer-based model, on English CC100~\citep{conneau-etal-2020-unsupervised}.
    }
    \item {\color{magenta}Language Adapt}: The {\color{magenta}token embedding layer} is finetuned using unlabelled data in the new language, while the {\color{gray}transformer body} is frozen.
    \item {\color{orange}Task Adapt}: The {\color{orange}transformer body} is finetuned using downstream task data in English, while the {\color{gray}token embedding} layer is frozen.
    \item {\color{teal}Assemble}: The final model is assembled by taking the adapted token embedding layer from stage 2 and the adapted transformer body from stage 3. \todo{color blind friendly}
\end{enumerate}
\begin{figure*}
    \centering
    \includegraphics[width=1.0\columnwidth]{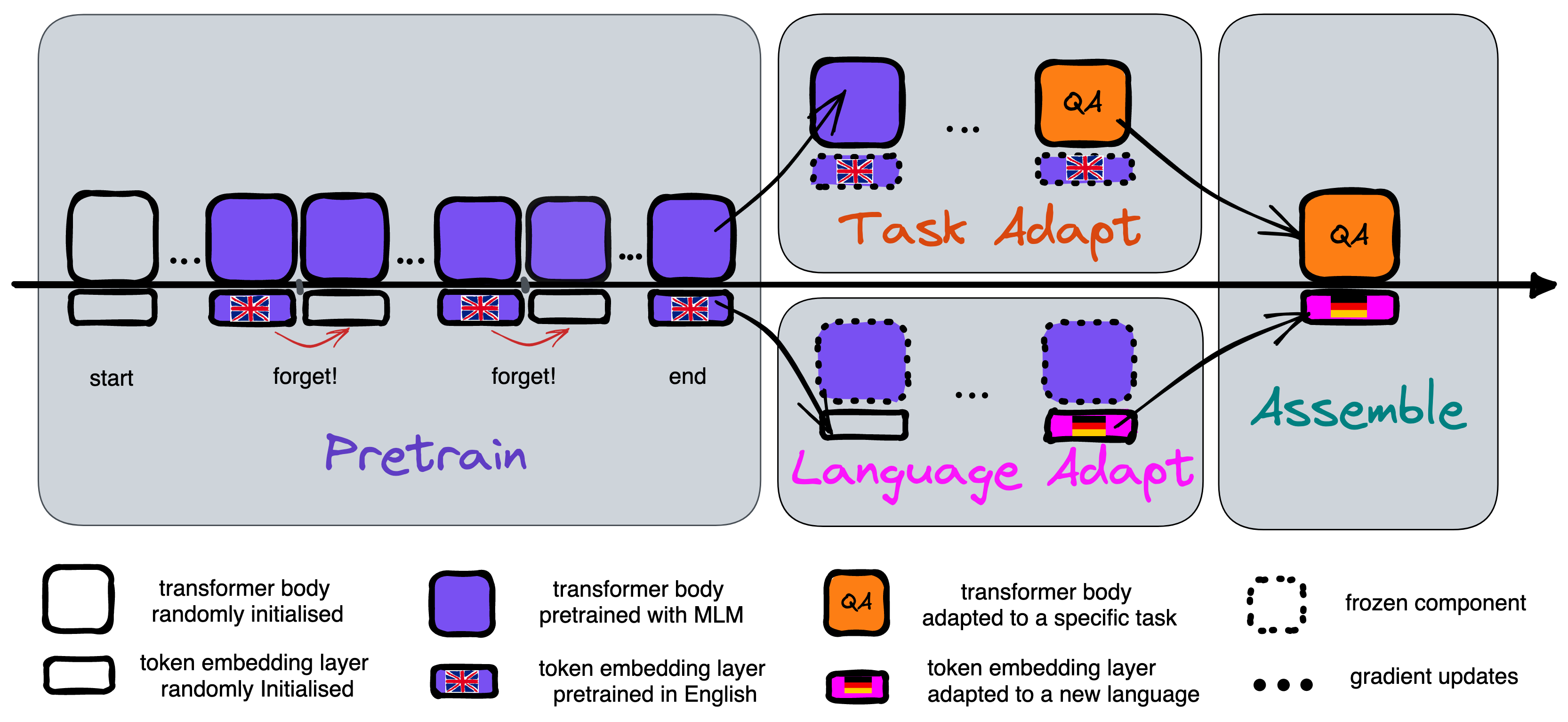}
    \caption{Unsupervised zero-shot cross-lingual transfer. \textbf{Left}: in the pretrain stage, we compare standard pretraining with forgetting pretraining, where the token embeddings are actively forgotten at a regular interval while the transformer body is learned as the standard pretraining. \textbf{Middle}: the task adapt stage and the language adapt stage separately adapt the transformer body using English task data and the token embeddings using unlabeled data in the new language. \textbf{Right}: the assemble stage reassemble the adapted body and token embedding layer into a usable PLM.}
    \label{fig:XNLI_stages}
\end{figure*}
\subsection{On The Difficulty of Rewiring PLMs via Relearning the Token Embeddings}
\label{sec:rewire_hard}
While the above procedure~\citep{artetxe2020cross} offers a general framework for rewiring a monolingual PLM with unlabelled data in the new language, it is unclear how efficient such rewiring can be, including both sample efficiency and computational efficiency.
\begin{wrapfigure}{L}{0.618\textwidth}
  \begin{center}
    \includegraphics[width=0.55\textwidth]{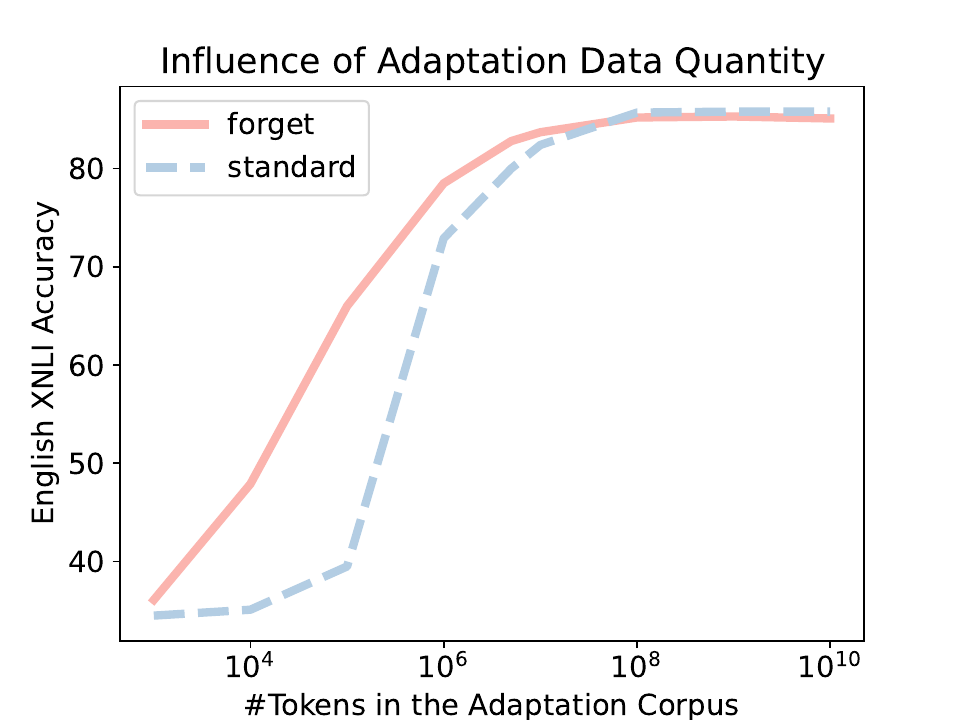}
  \end{center}
    \caption{The rewiring performance for standard PLMs (blue dashed line) drops drastically if the adaptation tokens $\leq 10\text{M}$.}
    \label{fig:relearn_en_vary}
\end{wrapfigure}
To better understand the difficulty of rewiring PLMs via relearning the token embeddings, we design an experiment where we relearn the token embedding layer using varying amounts of adaptation data. For illustration purpose, we pick English as the pseudo ``adaptation language'' because the English dataset is large enough to bootstrap a series of sub-datasets with varying quantity. We create sub-datasets with $[1\text{K}, 10\text{K}, 100\text{K}, 1\text{M}, 5\text{M}, 10\text{M}, 100\text{M}, 1\text{B}, 10\text{B}]$ tokens and relearn the English embeddings while keeping the transformer body frozen.

The dashed blue line in \Cref{fig:relearn_en_vary} summarizes the influence of the adaptation data quantity on the quality of the rewired PLMs (relearned embeddings assembled with the English NLI task body). We can see that the standard PLMs are easy to rewire if there is enough adaptation data. However if the adaptation corpus contains fewer than $10$ million tokens, the performance of the rewired standard PLMs (the blue dashed line in the figure) drops drastically as the adaptation data quantity goes down, from near $80$ to around $35$, a random-guessing level for NLI tasks. This motivates us to create more rewirable PLMs, i.e. PLMs with more plasticity so that the rewiring process can be faster and consume less data.

\section{Pretrain Easy-to-Rewire PLMs via Active Forgetting}
Recent works have shown that incorporating forgetting through iterative weights resetting can increase the ``plasticity'' of neural networks, enabling them to learn from small data and generalize better to unseen data in supervised learning~\citep{alabdulmohsin2021impact,taha2021knowledge,zhoufortuitous}. Building on these efforts, we study if we can bring such forgetting into the {\color{violet}pretrain} stage so that the resulting PLM would have more plasticity, allowing easier rewiring to new languages.

\paragraph{Our Hypothesis.} In effect, when \citet{artetxe2020cross} relearned the token embedding layer, the reinitialization of the embeddings can be seen as forgetting applied \textit{once} at the start of the {\color{magenta}language adapt} stage. However, the PLM~(specifically the transformer body) has never encountered forgetting before this stage and may struggle to handle this new situation. 
Without early exposure to forgetting, the PLM might suffer from slow recovery caused by forgetting before eventually benefiting from it.
The learning of a new lexical embedding layer in a PLM henceforth consumes lots of data in new languages along with long training horizons as shown in \Cref{sec:rewire_hard}.
In this paper, to ensure swift learning of the new languages with both high sample efficiency and convergence rate, we argue that the PLM must be exposed to forgetting during pretraining, allowing itself to maximize the positive impact of forgetting and minimizing the cost of recovery.

\begin{figure}
\includegraphics[scale=0.275,left]{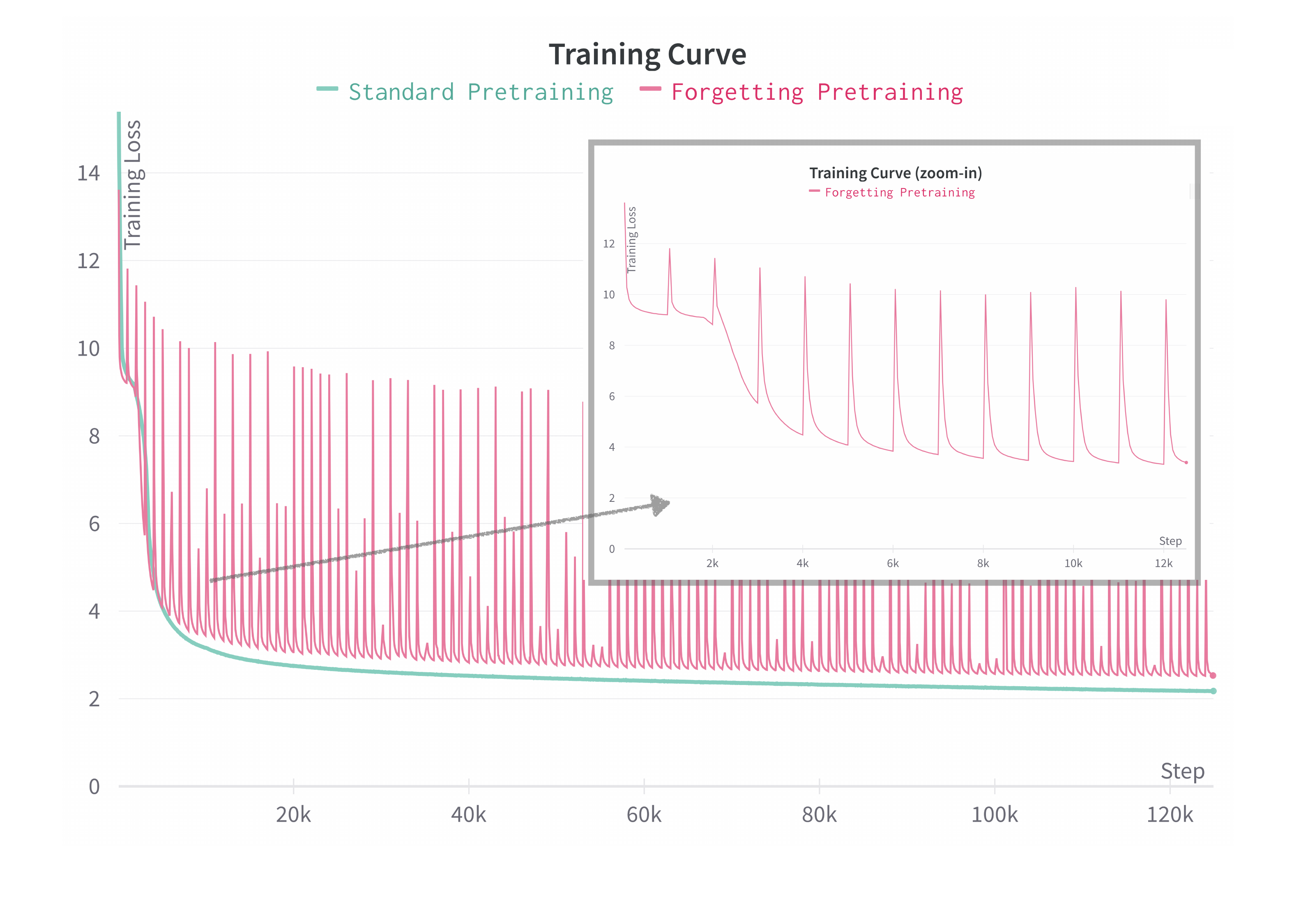}
\caption{Pretraining loss curves of forgetting and standard language models. The forgetting mechanism brings an episodic pattern into the loss curve: every embedding forgetting produces a loss spike, from which the model learn to recover. Through such repeats of forget-relearn, the model gets used to learn new embeddings from scratch.}
\label{fig:loss}
\end{figure}

\paragraph{Our Method.} With this hypothesis in mind, we propose to add an \emph{active forgetting} mechanism to the pretraining procedure, which resets the token embedding layer periodically as described in \Cref{algo:forgetting}.
Concretely, the forgetting mechanism operates by intentionally clearing the weights of the embedding layer, which stores the static representations for all tokens, and reinitializing them to a new set of random values every $K$ gradient updates.
Since pretraining involves advanced training strategies, like optimizers with states and learning rate schedulers, we also reset them together with the token embedding layer. 
We refer to language models pretrained with such active forgetting mechanism as \emph{forgetting PLMs}, in contrast to \emph{standard PLMs} which are pretrained in a standard way. 
The pretraining loss curve of a forgetting PLM is episodic (\Cref{fig:loss}), like in reinforcement learning or meta-learning. This episodic learning demonstrates that the active forgetting mechanism can introduce diversity without requiring actual new data. Each forgetting event kind of ``branches out'' a novel environment for the model to explore, as if initiating a new episode of learning.

\paragraph{Research Questions.} To further examine the proposed forgetting mechanism, we compare \emph{forgetting PLMs} and \emph{standard PLMs} on sample efficiency and convergence speed during {\color{magenta}language adapt}, two key aspects of model plasticity.
Our research investigates:
\begin{itemize}
    \item RQ1: Real-world low-resource languages often have scarce data for adapting models. Does pretraining with active forgetting impart enough plasticity to forgetting PLMs, enabling them to learn new languages even with such limited data?
    \item RQ2: Deploying PLMs frequently encounters computational limitations. Endowed with more plasticity, can forgetting PLMs reduce adaptation time for such low-compute scenarios?
    \item RQ3: New languages may be very similar or different from pretraining languages. Does this similarity/difference impact the relative benefit of forgetting PLMs over standard PLMs? 
\end{itemize}

\begin{algorithm*}\small
\begin{minipage}[c]{\textwidth}
\begin{algorithmic}[1]
\REQUIRE
$K$, interval between two consecutive forgetting; 
$n_{\text{body}/\text{emb}}$, current effective number of updates for the body or the token embedding layer; 
$\alpha_{\text{body}/\text{emb}}$, current learning rate for the body or the token embedding layer; 
$P^{n}_{\text{body}/\text{emb}}$, parameters after the $n^{\text{th}}$ update  for the body or the token embedding layer; 
$O^{n}_{\text{body}/\text{emb}}$, optimizer states after the $n^{\text{th}}$ update for the body or the token embedding layer; 
$\Theta$, randomly initialized embedding parameters, each element drawn from $\mathcal{N}(0, 0.02)$;
$f$, the function that computes the gradients w.r.t the parameters using the sampled data;
$g$, the function that updates the parameters based on the gradients (e.g. one step in Adam optimizer) 
$s$, the function that updates the learning rate (e.g. one step in the polynomial learning rate scheduler). 
\ENSURE the updated parameters and optimizer states $P^{(n)} = \{P^{(n)}_{\text{emb}}, P_{\text{body}}^{(n)}\}$, $O^{(n)} = \{O^{(n)}_{\text{emb}}, O_{\text{body}}^{(n)}\}$. 
\STATE $n_{\text{emb}} = n \text{ mod } K$
\STATE $\alpha_{\text{body}} = s(n_{\text{body}})$ \COMMENT{adjust learning rate based on $n$} 
\STATE $\alpha_{\text{emb}} = s(n_{\text{emb}})$
\STATE $G^{(n)} = f(P^{(n-1)}, \cdot)$ \COMMENT{compute all gradients}
\STATE $P_{\text{body}}^{(n)}, o_{\text{body}}^{(n)} = g(G_{\text{body}}^{(n)}, P_{\text{body}}^{(n-1)}, o_{\text{body}}^{(n-1)},\alpha_{\text{body}}, n)$ \COMMENT{update the transformer body}
\IF{$n_{\text{emb}} == 0$} 
 \STATE $P_{\text{emb}}^{(n)} = \Theta$ \COMMENT{periodical reset token embeddings and the relevant optimizer states}
\STATE $o_{\text{emb}}^{(n-1)} = 0$
\ENDIF
\STATE $P_{\text{emb}}^{(n)}, o_{\text{emb}}^{(n)} = g(G_{\text{emb}}^{(n)}, P_{\text{emb}}^{(n-1)}, o_{\text{emb}}^{(n-1)}, \alpha_{\text{emb}}, n_{\text{emb}})$ \COMMENT{update the token embeddings}
\end{algorithmic} 
\caption{Active forgetting mechanism. The token embedding layer is reset every $K$ updates.}
\label{algo:forgetting}
\end{minipage}
\end{algorithm*}

\section{Evaluate Forgetting PLMs for Unsupervised Cross-Lingual Transfer}
To evaluate the effectiveness of forgetting PLMs and address RQ1-RQ3, we conduct experiments on several cross-lingual transfer benchmarks.
\subsection{Experimental Setup} 
In our work, we closely follow the setup in \citet{artetxe2020cross} and \citet{marchisio2022mini}. Our {\color{violet}pretraining} model is RoBERTa-base, a standard $12$-layer transformer-based language model. We trained language-specific sentencepiece tokenizers~\citep{kudo2018sentencepiece} with a vocabulary size of $50$K over the corresponding data subsets in CC100. The model was pretrained with the English subset of the CC-100 dataset. The pretraining process consists of $125$K updates, with a batch size of 2048.
We used a learning rate scheduler with linear decay and an initial learning rate of $7\text{e}-4$, with $10$K warm-up updates. Checkpoints were saved every $500$ updates and we always choose the last pretraining checkpoint where possible for optimal performance. For forgetting pretraining, we chose the checkpoint corresponding to the best validation perplexity since the last checkpoint might have token embeddings reset. We set the frequency of forgetting $\text{K}=1000$  and used a clip-norm of $0.5.$

During the {\color{magenta}language adapt} stage, we kept most of the hyper-parameters the same as for pretraining.
We finetuned the token embedding layer while keeping the others frozen, as described in \Cref{sec:bg}. 
Note that \emph{no} forgetting happens during this stage because we want the models to learn the new languages as well as possible.
In the {\color{orange}task adapt} stage, both models were finetuned for $10$ epochs on the English task data, specifically MultiNLI~\citep{williams2018broad} for the NLI task and SQUAD~\cite{rajpurkar2016squad} for the QA task. 
After the {\color{teal}assemble} stage, we evaluate the zero-shot performance of the assembled model on XNLI~\citep{conneau-etal-2018-xnli}, a cross-lingual NLI task, along with XQuAD~\citep{artetxe2020cross} and MLQA~\citep{lewis2020mlqa}, two cross-lingual QA tasks. We report the NLI accuracy and QA F1 on the test sets. 

Our experiments were implemented using fairseq~\citep{ott2019fairseq}. The pretraining and language adaptation experiments were conducted on $32$ Tesla V100 GPUs (each with $32$ GB memory) and took approximately $24$-$36$ hours to complete. The time taken for both stages were quite close to each other even though the latter only involved tuning the embeddings. This demonstrates the importance of reducing the computational cost of the language adaptation stage.

Differing from prior work \citep{artetxe2020cross,marchisio2022mini}, we focus on  {\color{magenta}language adapt} in low-data regimes. We simulate low-resources scenarios by limiting the adaptation data for each downstream language to only $5$M subword tokens from CC100.
This is in contrast with conventional setups, where all the tokens in the corresponding languages in CC100 are used for language adaptation. As \Cref{tab:data_stat} shows, such setups consume several orders of magnitude more data than our $5$M-token setup; for instance, the Swahili CC100 subset contains $345$M tokens, roughly $69$ times larger than our corpus, and the Russian subset contains $34.9$B tokens, roughly $6,980$ times larger.
Therefore, PLMs that can successfully learn new languages with rich data under traditional setups may struggle to do so with our limited $5$M-token corpus.

\begin{table*}[t]
    \centering
    \resizebox{0.85 \textwidth}{!}{%
\begin{tabular}{l|cccccccc}
    \toprule
    \textbf{Method} &     \textbf{vi} &     \textbf{sw} &     \textbf{es} &     \textbf{bg} &      \textbf{de} &      \textbf{fr} &      \textbf{el} &      \textbf{ru} \\
    \midrule
    Standard   & $\mathbf{65.8}$ &   $55.6$          &   $68.0$          &   $65.5$          &    $62.2$          &    $63.5$          &    $63.1$          &    $56.9$ \\
    Forgetting &  $62.8$         &   $\mathbf{59.5}$ &   $\mathbf{74.0}$ &   $\mathbf{71.7}$ &    $\mathbf{68.5}$ &    $\mathbf{71.2}$ &    $\mathbf{70.8}$ &    $\mathbf{65.8}$ \\
    \midrule
    Relative Gain     &  $-4.6\%$ &  $+7.0\%$ &  $+8.8\%$ &  $+9.5\%$ &  $+10.1\%$ &  $+12.1\%$ &  $+12.2\%$ &  $+15.6\%$ \\
    \bottomrule
\end{tabular}

    }
    \caption{Accuracy comparison of forgetting and standard PLMs on XNLI (table continues).}
    \label{tab:xnli_test_1}
\end{table*}
\begin{table*}[t]
    \centering
    \resizebox{0.85 \textwidth}{!}{%
\begin{tabular}{l|cccccc|c|c}
    \toprule
    \textbf{Method} &     \textbf{zh} &      \textbf{ur} &      \textbf{hi} &      \textbf{tr} &      \textbf{ar} &      \textbf{th} &     \textbf{Avg} & \textbf{en} \\
    \midrule
Standard &    $53.2$ &    $36.8$ &    $39.7$ &    $38.9$ &    $41.2$ &    $35.3$ &    $53.3$ & $\mathbf{86.1}$\\
Forgetting   &   $\mathbf{63.5}$ &   $\mathbf{45.8}$ &   $\mathbf{52.9}$ &   $\mathbf{52.7}$ &   $\mathbf{59.5}$ &   $\mathbf{59.7}$ &   $\mathbf{62.7}$ & $85.1$\\

    \midrule
    Relative Gain    & $+19.4\%$ & $+24.5\%$ & $+33.2\%$ & $+35.5\%$ & $+44.4\%$ & $+69.1\%$ &$+21.2\%$ &$-1.2\%$ \\
    \bottomrule
\end{tabular}

    }
    \caption{Accuracy comparison of forgetting and standard PLMs on XNLI (table continued). On average, forgetting PLMs achieve a $21.2\%$ relative gain in accuracy compared to standard PLMs across the languages tested, where averaged relative gain $ = \frac{\sum_{x \in \{\text{languages}\}} \text{Relative Gain of } x}{\# \text{Languages}}$.}
    \label{tab:xnli_test_2}
\end{table*}

\begin{table}[t]
    \centering
    \resizebox{0.85 \textwidth}{!}{%

\begin{tabular}{l|cccccc|c|c}
    \toprule
    \textbf{Method} &      \textbf{es} &      \textbf{vi}  &      \textbf{de} &      \textbf{zh} &      \textbf{hi} &      \textbf{ar} &     \textbf{Avg} & \textbf{en} \\
    \midrule
    Standard   &    $49.4$ &    $38.3$ &    $45.3$ &    $34.1$ &    $17.7$ &    $20.8$ &    $34.3$ & $\mathbf{78.9}$\\
    Forgetting     &    $\mathbf{55.3}$ &    $\mathbf{45.0}$ &    $\mathbf{53.4}$ &    $\mathbf{43.0}$ &    $\mathbf{28.8}$ &    $\mathbf{34.7}$ &    $\mathbf{43.4}$ & $78.3$ \\
    \midrule
    Relative Gain &  $+12.0\%$ &  $+17.6\%$ &  $+17.8\%$ &  $+26.2\%$ &  $+62.5\%$ &  $+67.0\%$ &  $+33.8\%$ &$-0.8\%$\\
    \bottomrule
\end{tabular}

   }
   \vspace{2mm}
    \caption{F1-score comparison of forgetting and standard PLMs on MLQA. On average, forgetting PLMs achieve a $33.8\%$ relative gain in F1 compared to standard PLMs across the languages tested, where averaged relative gain $ = \frac{\sum_{x \in \{\text{languages}\}} \text{Relative Gain of } x}{\# \text{Languages}}$.}
    \label{tab:mlqa_test}
\end{table}

\begin{table*}[]
    \centering
    \resizebox{\textwidth}{!}{%
\begin{tabular}{l|cccccccccc|c}
\toprule
\textbf{Method} & \textbf{vi} & \textbf{es} & \textbf{ru} & \textbf{de} & \textbf{el} & \textbf{zh} & \textbf{hi} & \textbf{ar} & \textbf{th} & \textbf{tr} & \textbf{Avg} \\
\midrule
Standard & $49.7$ & $57.7$ & $49.4$ & $50.9$ & $48.5$ & $32.4$ & $21.4$ & $22.2$ & $15.4$ & $13.0$ & $36.1$ \\
Forgetting & $\mathbf{52.9}$ & $\mathbf{64.6}$ & $\mathbf{56.5}$ & $\mathbf{60.9}$ & $\mathbf{59.9}$ & $\mathbf{43.7}$ & $\mathbf{33.3}$ & $\mathbf{38.7}$ & $\mathbf{38.4}$ & $\mathbf{41.4}$ & $\mathbf{49.0}$ \\
\midrule
Relative Gain & $+6.4\%$ & $+12.0\%$ & $+14.5\%$ & $+19.7\%$ & $+23.6\%$ & $+34.6\%$ & $+55.8\%$ &$+74.2\%$ &$+149.7\%$ &$+218.8\%$ &$+60.9\%$ \\
\bottomrule
\end{tabular}
}    
    \caption{F1-score comparison of forgetting and standard PLMs on XQuAD. On average, forgetting PLMs achieve a $60.9\%$ relative gain in F1 compared to standard PLMs across the languages tested, where averaged relative gain $ = \frac{\sum_{x \in \{\text{languages}\}} \text{Relative Gain of } x}{\# \text{Languages}}$.}
    \label{tab:xquad_test}
\end{table*}

\subsection{Forgetting PLMs Work Better in Low-Data Regimes (RQ1)} 
Standard PLMs struggle in low-data language adaptation, dropping from $86.1$ English NLI accuracy to just $53.3$ average accuracy on XNLI with limited 5M token adaptation data. Compared to prior work which uses full data from Wikipedia~\citep{artetxe2020cross} or from CC100~\citep{marchisio2022mini}, the average accuracy on XNLI drops about $18\%$ (from $66.8$/$66.3$ to $53.3$). This indicates standard PLMs are not coping well with the low-data regime.
In contrast, forgetting PLMs achieve decent $62.7$ average XNLI accuracy, a {\color{blue}$+21.2\%$} relative gain over standard PLMs, as shown in \Cref{tab:xnli_test_2}.

Forgetting PLMs also outperform standard PLMs on MLQA and XQuAD, with average F1 relative gains of {\color{blue}$+33.8\%$} and {\color{blue}$+60.9\%$} across languages, as respectively demonstrated in \Cref{tab:mlqa_test} and \Cref{tab:xquad_test}. Across NLI and QA tasks, forgetting PLMs consistently surpass standard PLMs in low-data regimes.
Why do forgetting PLMs handle the low-data regime better? We hypothesize this is because forgetting PLMs are more robust to different embedding initializations. They encode more universal knowledge in the transformer body. Standard PLMs may encode more ``shortcut'' knowledge relying on certain embedding initializations. In low data, standard PLMs cannot adjust embeddings towards shortcut routes without access to enough data. Forgetting PLMs do not rely on shortcuts so perform better.

\subsection{Forgetting PLMs Learn New Languages with Fewer Parameter Updates (RQ2)}
\begin{figure*}
  \centering
  \begin{subfigure}{0.32\textwidth}
    \includegraphics[width=\textwidth]{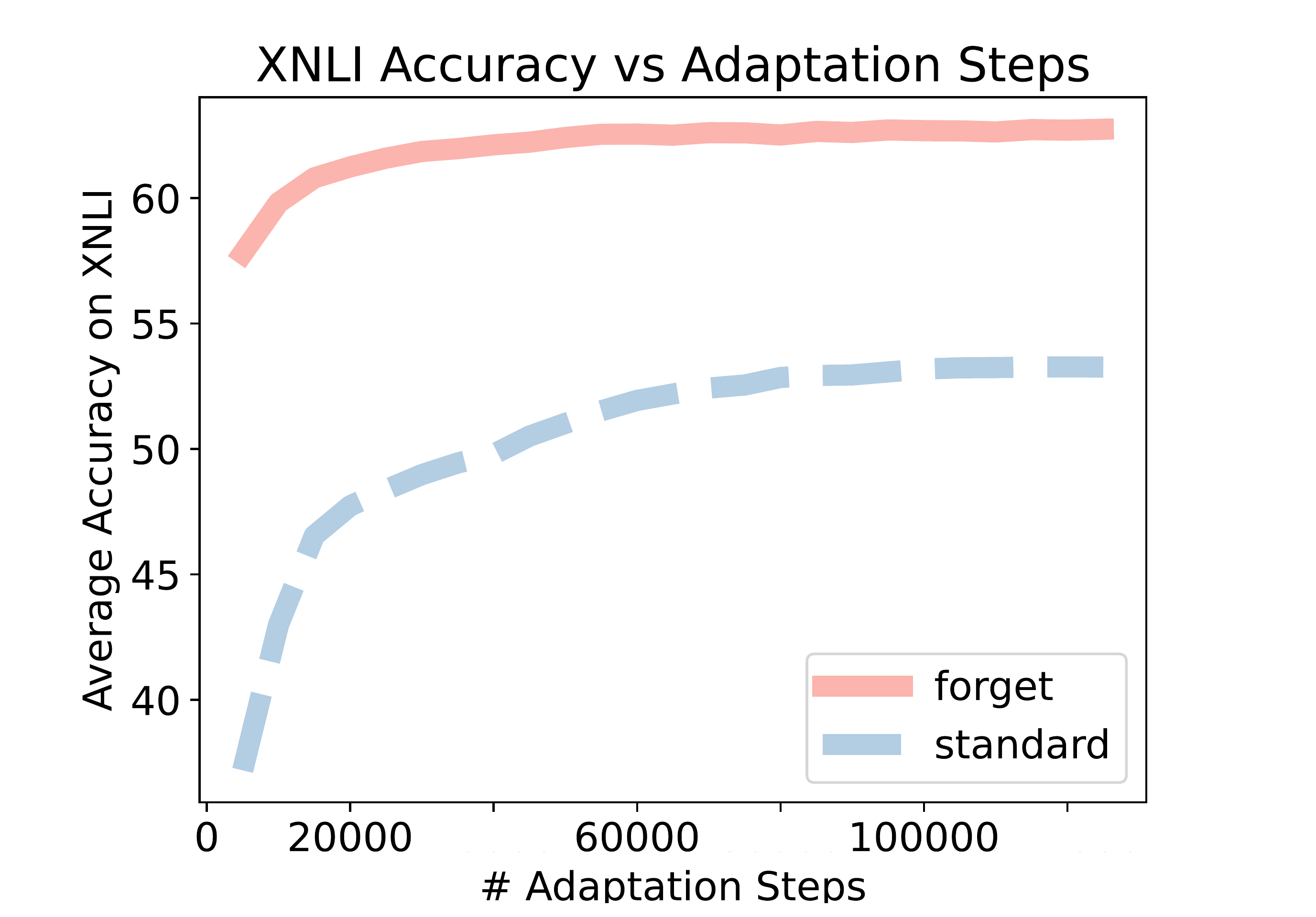}
  \end{subfigure}
\hfill
  \begin{subfigure}{0.32\textwidth}
    \includegraphics[width=\textwidth]{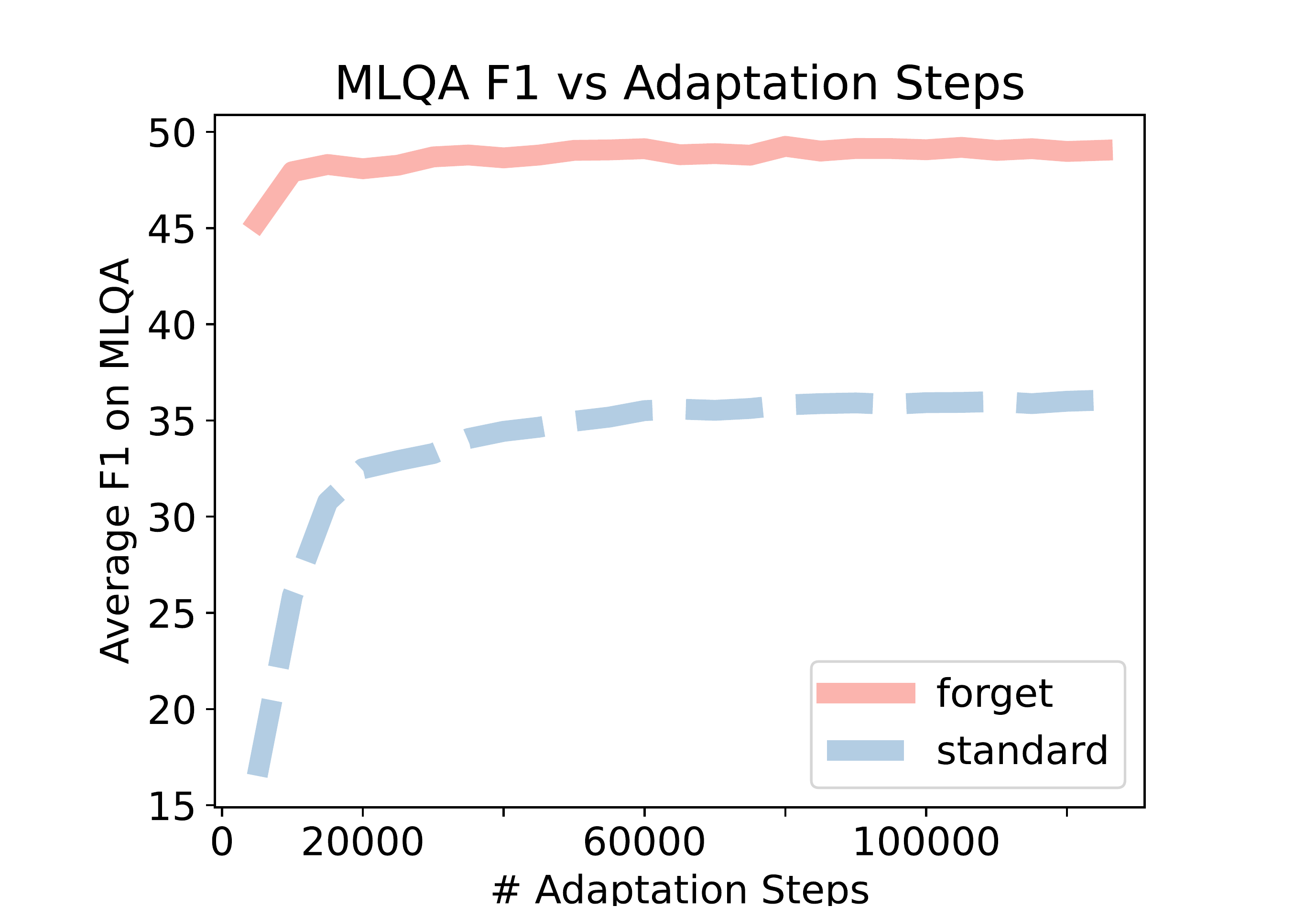}
  \end{subfigure}
 \hfill
  \begin{subfigure}{0.32\textwidth}
    \includegraphics[width=\textwidth]{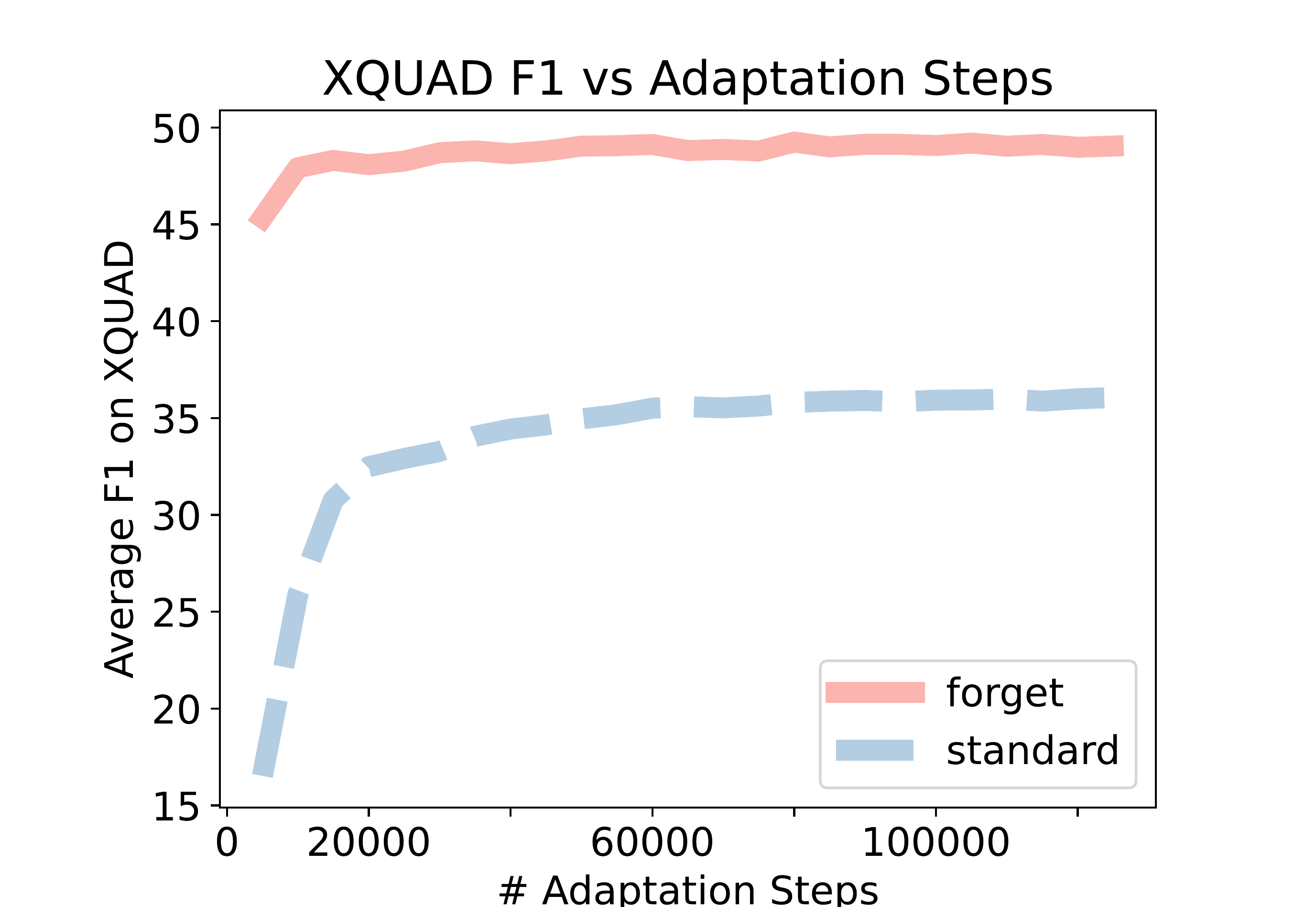}
  \end{subfigure}
  \vspace{1em}
  \begin{subfigure}{0.3\textwidth}
    \includegraphics[width=\textwidth]{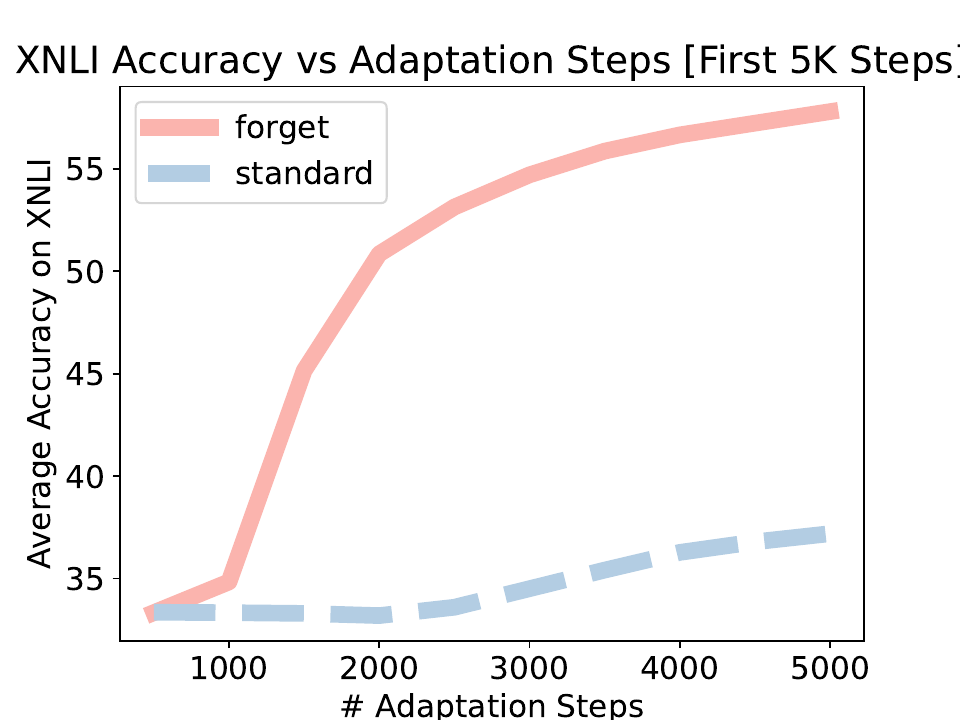}
  \end{subfigure}
  \hfill
  \begin{subfigure}{0.3\textwidth}
    \includegraphics[width=\textwidth]{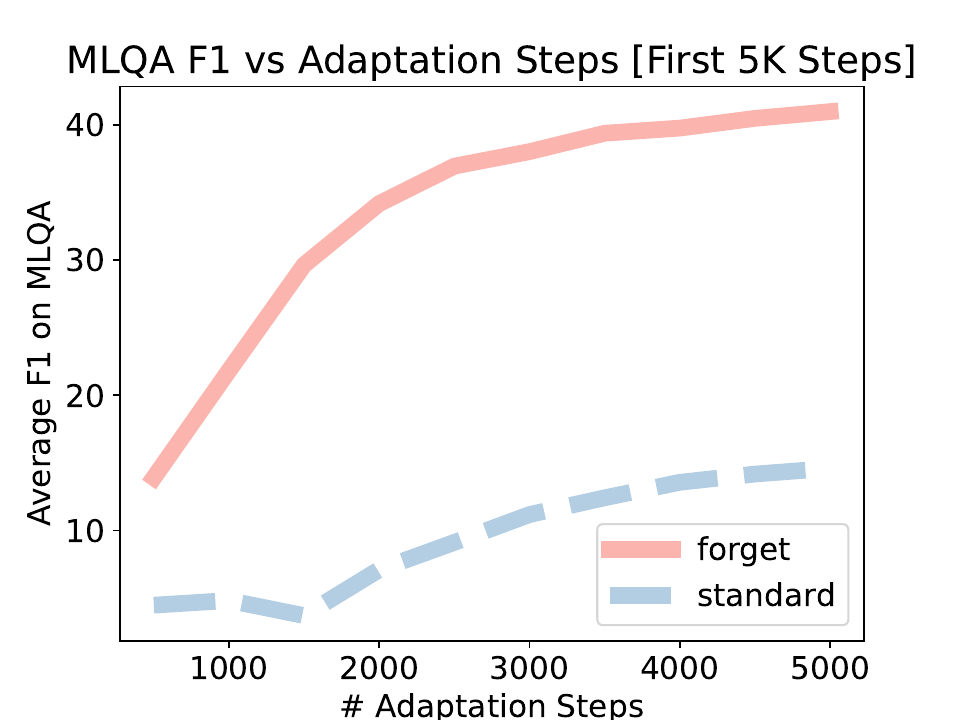}
  \end{subfigure}
  \hfill
  \begin{subfigure}{0.3\textwidth}
    \includegraphics[width=\textwidth]{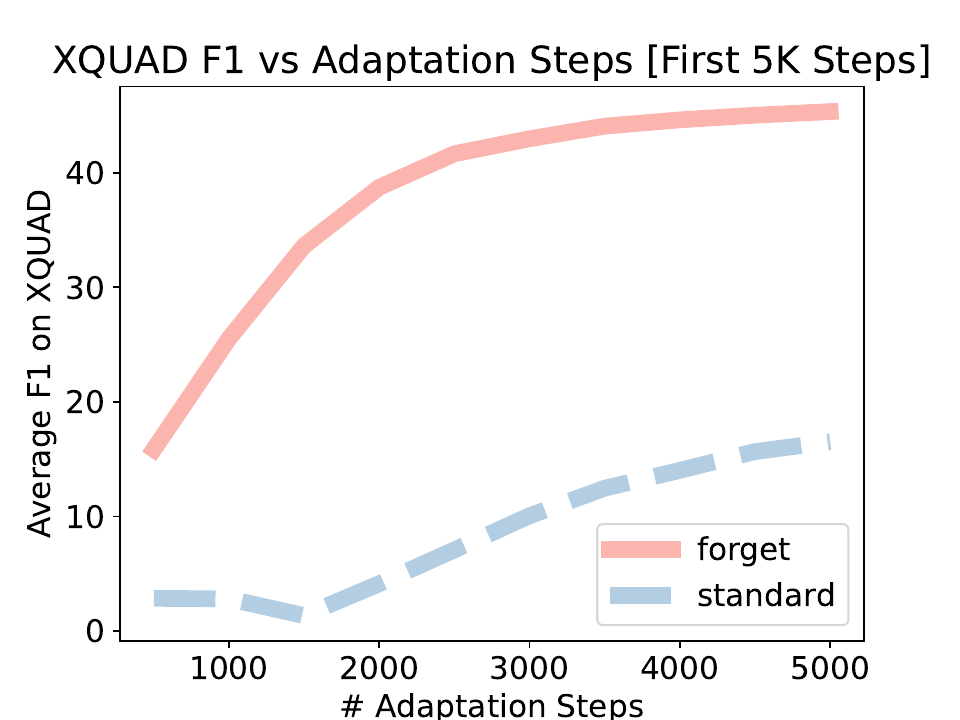}
  \end{subfigure}
  \caption{Adaptation curves on XNLI, MLQA, and XQuAD. Numbers aggregated across languages. The first row contains the full adaptation curves, which comprises $125\text{K}$ adaptation steps. The second row contains the zoom-in versions of curves for the first $5\text{K}$ adaptation steps. Forgetting PLMs converge faster than standard PLMs; for instance, on XQuAD (the last plot), forgetting PLMs reach $92\%$ of their final performance within $5\text{K}$ updates, while standard PLMs only reached $53\%$ of their final performance at that point.}
  \label{fig:adaptation_curve}
\end{figure*}
We are also interested in how quickly forgetting PLMs and standard PLMs can learn new languages. \Cref{fig:adaptation_curve} summarizes adaptation curves on XNLI, MLQA and XQuAD, with each point representing the averaged performance across all languages. 
In just $5\text{K}$ steps ($4\%$ of full adaptation), forgetting PLMs reach $57.8$ accuracy on XNLI while standard PLMs struggle at random guessing levels of $37.2$. Similar trends hold for MLQA and XQuAD. After $5\text{K}$ steps, forgetting PLMs achieve {\color{blue}$92\%$} of their full performance on XQuAD versus just {\color{blue}$53\%$} for standard PLMs (see the last plot in \Cref{fig:adaptation_curve}).

Why do forgetting PLMs converge faster? We hypothesize it is because periodical embedding resetting forces the body to gradually locate itself on a particular manifold, where it can easily cooperate with new embeddings.
This makes the body encourage larger embedding updates when adapting to new languages.
Active forgetting simulates language switching during pretraining\footnote{Precisely, it simulates vocabulary swappings, causing drastic changes to the input of the body.} introducing diversity without new data. This allows faster adaptation to real new languages.

\subsection{Languages That Are Distant To English Benefit Most From Forgetting PLMs (RQ3)}
\begin{figure}
  \centering
  \begin{subfigure}[b]{0.48\textwidth}
  \includegraphics[width=\textwidth]{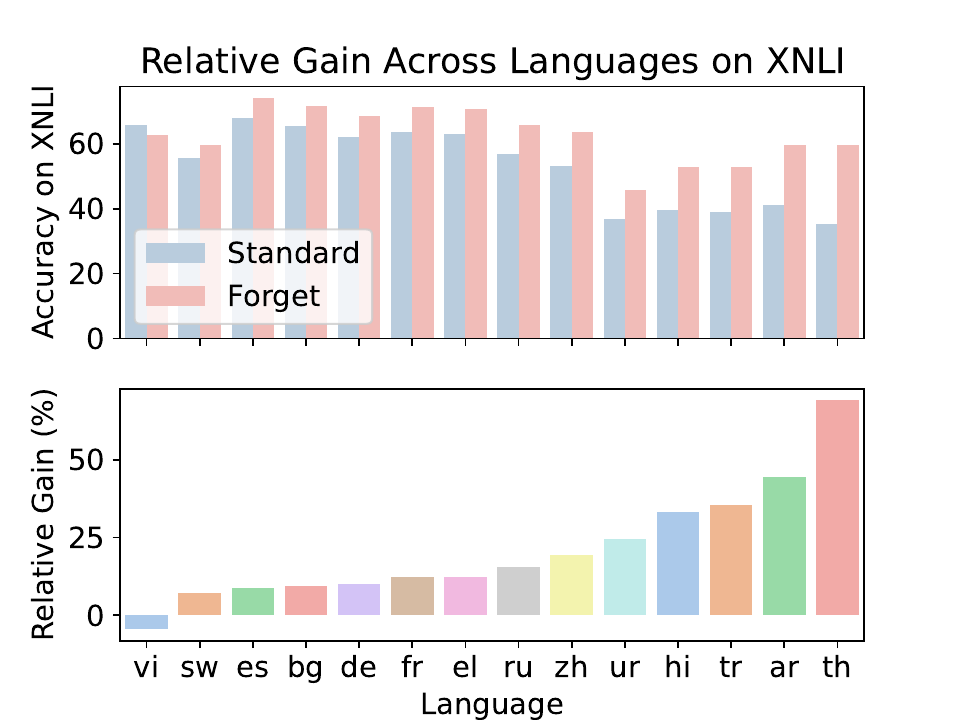}
  \end{subfigure}
  \begin{subfigure}[b]{0.48\textwidth}
  \includegraphics[width=\textwidth]{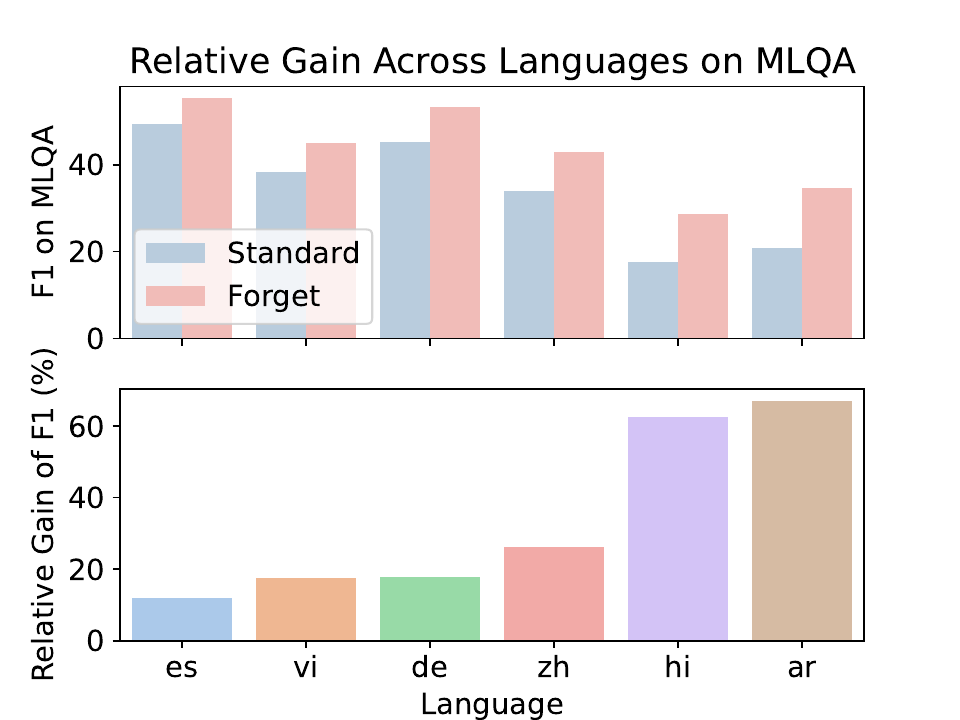}
  \end{subfigure}
  \caption{Relative gains of forgetting PLMs over standard PLMs across languages. Forgetting yields substantial relative gains for languages like Arabic, Hindi, Thai, Turkish, and Urdu. However, for languages closely related to English, such as German, the relative gains from forgetting are modest.}
  \label{fig:relative_gain}
\end{figure}

\begin{figure}
    \centering
    \includegraphics[trim={ 0cm 0cm 0cm 0cm},clip,width=0.9\textwidth]{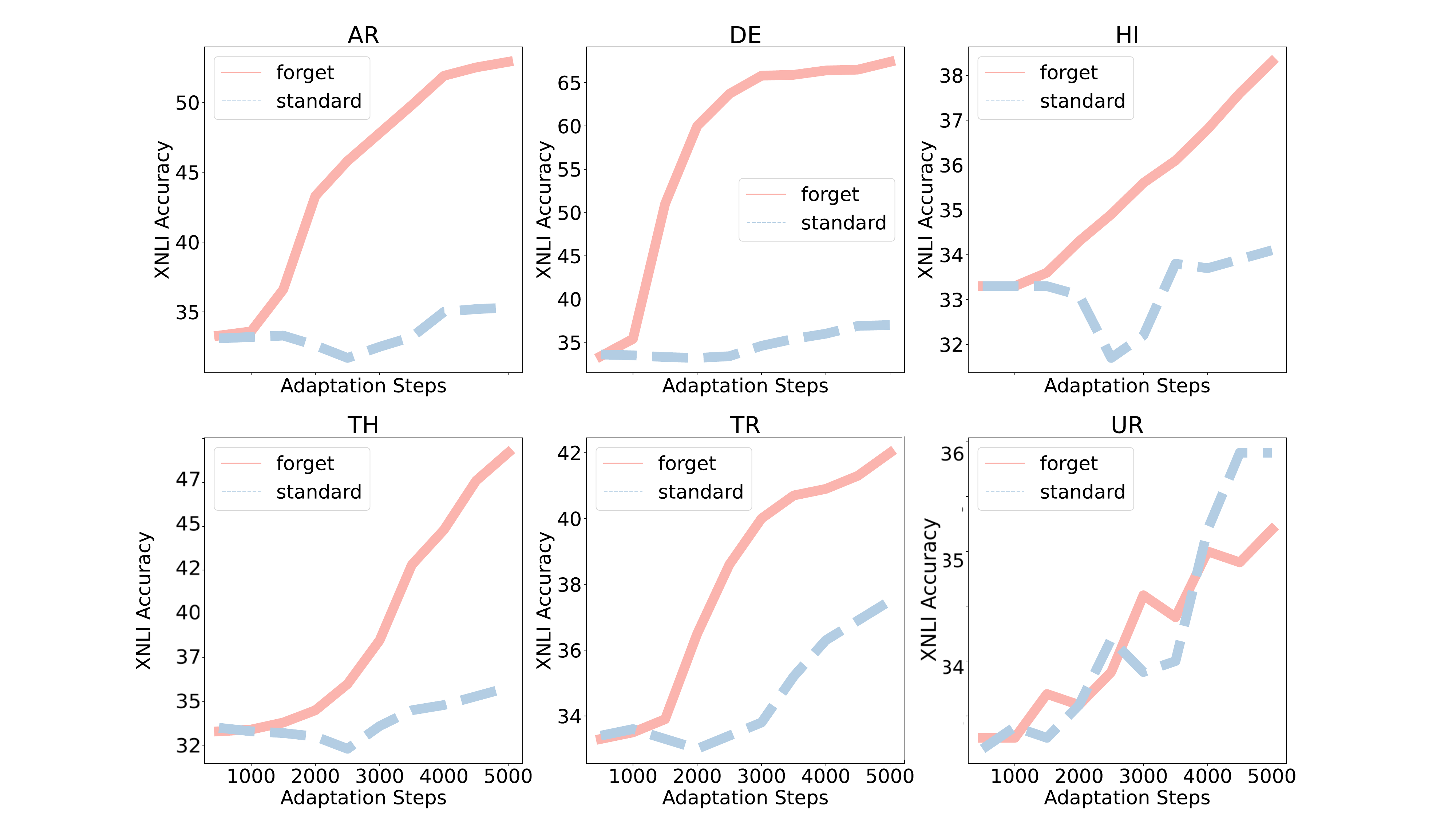}
    \caption{Adaptation curves on XNLI within $5\text{K}$ updates for individual languages: Bulgaria, Greek, Spanish, French, Russian, Swahili, Vietnamese and Chinese. For all languages except Urdu, the forgetting PLMs converge faster than the standard PLMs during the language adaptation stage.}
    \label{fig:5M_5k}
\end{figure}

Up to this point, we have primarily presented the averaged performance. In this section, we delve into a detailed comparison of language-specific performance between forgetting PLMs and standard PLMs on XNLI, MLQA, and XQuAD. To gain a deeper insight into which languages benefit the most from the use of forgetting, we present the relative performance changes in \Cref{fig:relative_gain} for XNLI and MLQA. The results for XQuAD can be found in \Cref{fig:relative_gain_xquad} in the appendix.
Across the spectrum of languages (\Cref{tab:lang_stat}), we observe that forgetting provides greater benefits for languages distant to the pretraining language (English) in terms of language family, script and morphology. Specifically, forgetting brings large relative gains for languages such as \textit{Arabic}, \textit{Hindi}, \textit{Thai}, \textit{Turkish}, and \textit{Urdu} compared to closer languages like \textit{German}. Script seems important - forgetting helps Vietnamese and Swahili less despite their distance from English, likely due to the shared Latin script. 
Examining adaptation curves within the first $5$K steps, forgetting PLMs reach substantially superior performance over standard PLMs for almost all languages except Urdu, while standard PLMs struggle at random guess levels (see \Cref{fig:5M_5k} and \Cref{sec:appendix_results_convergence}). This demonstrates forgetting PLMs' ability to efficiently adapt to new languages, particularly dissimilar ones, in low-data settings.

\section{Related Work}
\subsection{Forgetting and its Positive Roles} 
The common perception of forgetting is that it implies weak memory and a loss of acquired knowledge, thus it is often regarded as a sign of \emph{un-intelligence} or an undesirable property. In neural networks, \emph{catastrophic forgetting}~\citep{MCCLOSKEY1989109,ratcliff1990connectionist,kirkpatrick2017overcoming} is portrayed as a forgetting phenomenon where neural networks lose the ability to predict old patterns after new inputs alter their weights. Forgetting in this context has negative consequences, as the new knowledge overwrites the old.
Plenty of prior research strives to overcome catastrophic forgetting and enable continual learning~\citep{schmidhuber2013powerplay, kirkpatrick2017overcoming, lopez2017gradient, shin2017continual, schwarz2018progress, mallya2018packnet, parisi2019continual, rolnick2019experience, beaulieu2020learning, veniat2020efficient, gaya2022building, khetarpal2022towards}.

Our work differs from the above ones in that our subject is \emph{intentional forgetting} rather than passive forgetting and its associated negative impact. To put it in another way, we seek to understand how forgetting -- if purposely incorporated as an active process during training -- might \emph{help} new learning. Similar positive roles of forgetting have been discussed in the literature. 
Specifically, ~\citet{pastotter2008oscillatory} demonstrate forgetting enhances the learning of new information by resetting the encoding process and holding the attention at high levels; ~\citet{levy2007inhibiting} show that it helps second language acquisition by inhibiting the native language; ~\citet{barrett2009role} find it promote the emergence of an optimal language by preventing partial success from reinforce sub-optimal practice.
\citet{simon2015} further suggests forgetting serves adaptive functions, helping people regulate emotions, acquiring knowledge and staying attuned to the context. 
More recently \citet{anderson2021} reviews evidence on active forgetting by prefrontal control and shows how it can adapt the memory to suit either emotional or cognitive goals. 

\subsection{Forgetting Via Partial Neural Weights Reset}
In neural networks, forgetting can be instantiated in many forms. A simple way is to reset subsets of parameters before the next round of learning. Iterations of such resetting have been shown to benefit generalization with low compute and low data for computer vision tasks~\citep{frankle2018lottery,alabdulmohsin2021impact,taha2021knowledge,ramkumar2023learn}. 
More recently, ~\citet{zhoufortuitous} demonstrate a similar forgetting strategy helps image classification and language emergence.
Closely linked to our method, ~\citet{chenrefactor} forget node embeddings in order to truncate infinite message-passing among nodes and thereby aid new graph reasoning with new nodes. Our work uses similar forgetting mechanism over token embeddings, improving new language reasoning with new tokens.
As far as we know, \textit{we are the first to bring forgetting into pretraining and demonstrate that forgetting pretraining boosts linguistic plasticity}.
A relevant thread in reinforcement learning (RL) research studies the plasticity loss phenomenon ~\citep{pmlr-v202-lyle23b, nikishin2023deep}. Recent work explores similar forgetting approaches to improve plasticity. ~\citet{igl2021transient} periodically reset the current policy by distilling it into a reinitialized network throughout training. Intuitively, this releases network capacity storing suboptimal policies and opens up space for the the yet-to-be-discovered optimal (final) policy. 
Simpler methods just reset an agent's last layers~\citep{nikishin2022primacy}, preventing overfitting to early experiences and \emph{primacy bias}. Resetting parameters also improves sample efficiency by allowing more updates per environment interaction ~\citep{d2022sample}.

\subsection{Making Pretrained Language Models Multilingual}
Pretraining on multilingual data makes PLMs multilingual~\citep{conneau-etal-2020-unsupervised} but has downsides like needing large multilingual corpus with appropriate mixing, potential interference among languages, and difficulty of covering all languages.  
Alternatively, the line of research on cross-lingual transfer makes PLMs multilingual by extending English-only PLMs to other languages. ~\citet{artetxe2020cross} demonstrate possibility of such extension by relearning the embedding layer with unsupervised data from the new language. ~\citet{marchisio2022mini} further increase computational efficiency using a mini-model proxy. ~\citet{liu2023delving} use a similar partial reset-reinit approach in vision-language settings. Approaches based on adapters and sparse finetuning have also been proposed~\citep{pfeiffer-etal-2020-mad,pfeiffer2022lifting,pfeiffer2021unks,ansell2022composable}. Adapters are bottleneck layers~(usually placed after the feedforward layers) that add extra capacity when adapting to a different task or language. Our proposed forgetting mechanism can be applied to adapter-based methods as we can allow forgetting to happen in the adapter layers. The current choice of forgetting embeddings keeps the architecture intact and incurs no additional hyperparameter tuning, allowing us to understand the fundamental capability of forgetting pretraining.

\section{Conclusion \& Future work}
\subsection{Conclusions}
While forgetting is usually perceived as negative, recent work points out that it can also be beneficial in certain cases, particularly for quickly learning new tasks, training in non-stationary environments~\citep{igl2021transient, nikishin2022primacy, d2022sample} and improving sample efficiency~\citep{taha2021knowledge,zhoufortuitous}. Joining this line of work, our paper demonstrates that forgetting techniques can improve pretrained language models by imbuing them with more linguistic plasticity. Specifically, our proposed \textit{active forgetting} mechanism can create PLMs that are easier to rewire for new lingual spaces. Experiments with RoBERTa show that models pretrained via active forgetting can better learn from small amounts of data while also enjoying faster convergence during language adaptation, particularly for languages distant from English. 

Going beyond language adaptation, we argue that PLMs with more plasticity are a promising direction for future research, as they allow easier adaptation to various tasks, domains, languages and can evolve faster as the real world changes.
Unlike symbolic methods, such as knowledge graphs, which can easily rewire a fact by modifying the corresponding knowledge triplet, current static PLMs are harder to rewire since changing one fact via updating model weights may disrupt multiple other facts without substantial post-hoc intervention.
Improving the rewirability via forgetting pretraining thus can be seen as one way of imbuing PLMs with similar benefits as symbolic methods (making the resulted model more controllable i.e. can be modified with tiny cost), complementing the line of post-hoc model editing research~\citep{mitchell2021fast,mitchell2022memory}.

\subsection{Limitations}
Our work uses the simplest forgetting approach - directly resetting embeddings to random initialization. Advanced techniques like gradually injecting noise could be explored. We focus on masked language modeling pretraining with language-specific tokenizers. Applying active forgetting to auto-regressive LMs, other pretraining methods (e.g. deberta pretraining~\citep{he2021deberta,he2021debertav3}), and various tokenization is promising future work. More broadly, current large language models need more plasticity to expand across tools, tasks, and domains. Our work takes an initial step, showing directly resetting embeddings can significantly improve model plasticity. Further research on more sophisticated forgetting techniques during pretraining could unlock additional gains.

On the theory front, potential connections can be made between forgetting and meta-learning~\citep{schaul2010metalearning, thrun2012learning, andrychowicz2016learning, finn2017model} since both attempt to learn solutions that can quickly adapt themselves to new inputs. 
Another possible theoretical explanation for why active forgetting works so well might be related to the flatness of the solution in the loss landscape~\citep{alabdulmohsin2021impact}. Flatter minima tend to enjoy better generalization~\citep{liu2022pretraining}. Thus, it might be worthwhile to study the flatness of the transformer body during the forgetting pretraining.

\newpage

\begin{ack}
We would like to thank our reviewers for their valuable suggestions. We are also grateful to those who engaged in interesting discussions during the project, including Pasquale Minervini, Xuanli He, Jiayi Wang, Yuxiang Wu, Hila Gonen, Dieuwke Hupkes, Fabio Petroni, Naila Murray, Alexis Thual, Nicola Cancedda, Yingchen Xu, and Hubert Jacob Banville. Yihong would like to express her thanks to the FAIR-UCL PhD program for generously funding her PhD. David Adelani acknowledges the support of DeepMind Academic Fellowship programme.
\end{ack}
\bibliography{ref}
\bibliographystyle{plainnat}

\newpage
\appendix
\section{Discussion on Experimental Setup A Low-Data Regime}
A common experimental setup for adapting to a target language is to use all the available data in that language from sources such as Wikipedia~\citep{artetxe2020cross,ansell2022composable} and CC$100$~\citep{marchisio2022mini}. In this setup, the numbers of tokens typically used for adapting each language might differ greatly, ranging from $13.9$M to $70.5$B, as summarized via \Cref{tab:lang_stat}.

Our work, however, investigates a different setup where we control the adaptation data to $5$ million tokens for each language. This can be highly relevant when studying generalisation to completely new languages, which require expanding the vocabulary. We acknowledge that dealing with real-world low-resources languages can be more challenging than such low-data setup used. And there are already rich work addressing low-resource issues: multilingual pretraining~\citep{conneau-etal-2020-unsupervised,pfeiffer2022lifting}, multilingual adapters~\citep{pfeiffer-etal-2020-mad,ansell2022composable}, multilingual adaptation~\citep{Tang2020MultilingualTW,alabi-etal-2022-adapting}, and multilingual regularization~\citep{pfeiffer2021unks}. Nevertheless, we would like to highlight the importance of our low-data regime. The challenge of ``low-resource'' involve multiple \textit{entangled factors}: the quality of the tokeniser, the amount of data, whether the script/language family is distant to the pretraining language(s) etc. Simulating a low-data regime allows us to control these factors and isolate the effects of the factor that we are interested in -- the amount of data in the new language. This factor is essential to our work as our research goal is plasticity i.e. rewiring model prediction with as little new information as possible. Simulating various amount of data in the new language allows us to compare model plasticity as shown in \Cref{fig:relearn_en_vary}, and thus contribute a clean piece of knowledge in the line of plasticity research~\citep{pmlr-v202-lyle23b,nikishin2023deep}.

\section{Discussion on Experimental Framework Choice for Studying Pretrained Language Models' Plasticity}
Our motivation is to improve language models' plasticity. Plasticity of neural networks have been studied in graph learning, computer vision and reinforcement learning~\citep{taha2021knowledge,chenrefactor,pmlr-v202-lyle23b,nikishin2023deep}, where forgetting-relearn methods show promise. Our goal is to study plasticity in the context of pretrained language models. We believe this is a emerging research direction and will thrive in the following years. However, translating the plasticity concept to the language model setting is not trivial due to the lack of clear experimental setups. We note that, despite the model differences, almost all language models begins with a token embedding layer. As often tied to a specific vocabulary, the token embedding layer limits the plasticity, preventing generalisation to a new vocabulary. This observation inspires us to explore the plasticity of language models by manipulating the token embedding layer. ~\citet{artetxe2020cross} draws our attention as it offers a nice experimental framework of only manipulating the token embedding layer for adapting between languages. 

\begin{table}
    \centering
    \resizebox{0.7\textwidth}{!}{
    \begin{tabular}{lllll}
\toprule
      Name & Code &          Family &      Script &       Morphology \\
\midrule
    Arabic &  ar &         Semitic & Arabic (Abjad) & Introflexive \\
    Bulgaria & bg & IE:Balto-Slavic & Cyrillic & Analytic\\
    German &  de &     IE:Germanic &    Latin &         Fusional \\
     Greek &  el &     IE:Hellenic &    Greek &         Fusional \\
   English &  en &     IE:Germanic &    Latin &         Analytic \\
    French &  fr &      IE:Romance &    Latin &         Fusional \\
     Hindi &  hi & IE:Indo-Iranian &     	
Devanagari  &         Fusional \\
   Russian &  ru & IE:Balto-Slavic &    	Cyrillic &         Fusional \\
   Spanish &  es &      IE:Romance &    Latin &         Fusional \\
   Swahili &  sw &      Niger-Congo:Bantu &    Latin &         Agglutinative \\
      Thai &  th &       Tai-Kadai &     Thai &         Analytic \\
   Turkish &  tr &          Turkic &    Latin &    Agglutinative \\
      Urdu &  ur & IE:Indo-Iranian &       Perso-Arabic &         Fusional \\
Vietnamese &  vi &   Austroasiatic &    Latin &         Analytic \\
   Chinese &  zh &    Sino-Tibetan & Chinese &         Analytic \\
\bottomrule
\end{tabular}}
 \vspace{2mm}
    \caption{Languages by family, script, and morphology.}
    \label{tab:lang_stat}
\end{table}
\begin{table}
    \centering
    \begin{tabular}{lrr}
    \toprule
        \textbf{Language} & \textbf{CC-100} & \textbf{Wikipedia} \\
    \midrule
        sw & $345$M & $13.9$M \\
        ur & $832$M & $41.5$M \\
        hi & $2.13$B & $54.6$M \\
        ar & $4.15$B & $337$M \\
        tr & $4.19$B & $157$M \\
        th & $6.09$B & $70.7$M \\
        el & $6.1$B & $148$M \\
        bg & $7.9$B & $134$M \\
        zh & $9.72$B & $584$M \\
        es & $11.6$B & $1.17$B \\
        fr & $13.3$B & $1.71$B \\
        de & $14.1$B & $2$B \\
        vi & $28.9$B & $300$M \\
        ru & $34.9$B & $1.25$B \\
       \textbf{en} & \textbf{$70.5$B} & \textbf{$4.25$B} \\
    \bottomrule
    \end{tabular}
    \caption{Numbers of tokens for different languages on the two multilingual corpus, CC-100 and Wikipedia, in ascending order of CC100. The English one is used as pretraining corpus while the others are used for language adaptation.}
    \label{tab:data_stat}
\end{table}
\section{More Results on Distant Languages Benefits More From Forgetting}
\begin{figure}
  \centering
  \begin{subfigure}[b]{0.618\textwidth}
  \includegraphics[width=\textwidth]{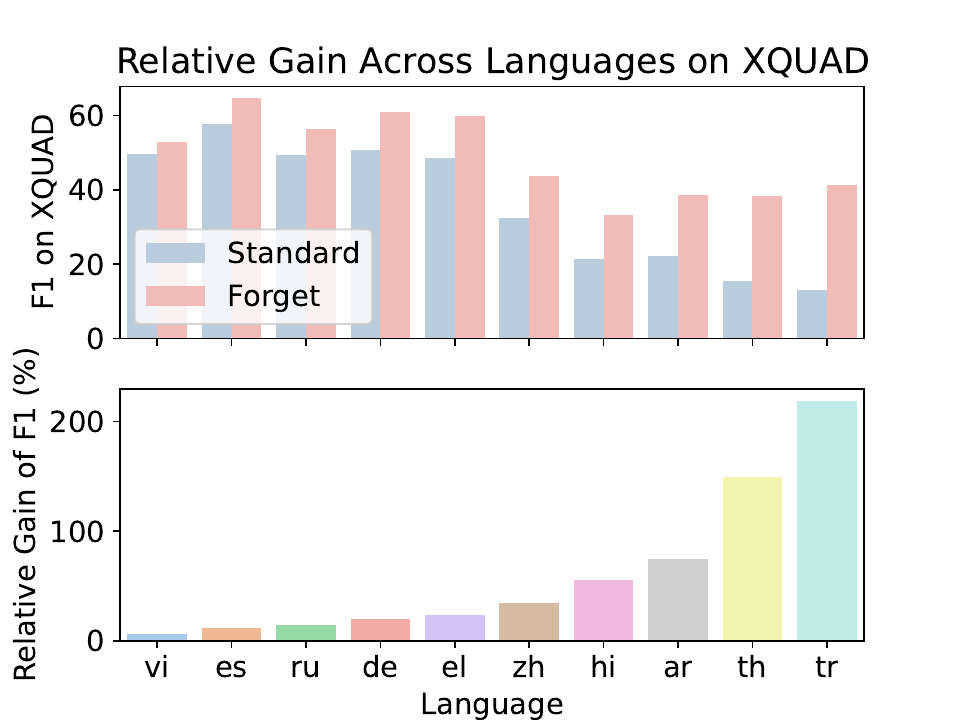}
  \end{subfigure}
  \caption{Relative gains of forgetting PLMs over standard PLMs across languages on XQuAD. Languages that are less related to the pretraining language (English), such as Turkish, Thai, Arabic, Hindi, benefit more from forgetting PLMs.}
  \label{fig:relative_gain_xquad}
\end{figure}

We are interested in how forgetting PLMs can improve adaptation to different languages. We compare the results of various languages on three benchmarks: XNLI, MLQA and XQuAD. We use \Cref{fig:relative_gain} and \Cref{fig:relative_gain_xquad} to illustrate the relative gains from active forgetting on each benchmark. We find that languages that are less related to the pretraining language, which in this case is English, benefit more from forgetting PLMs.

\section{More Results on Forgetting PLMs Converge Faster In The Language Adaptation Stage}
\label{sec:appendix_results_convergence}
\begin{figure}
    \centering
    \includegraphics[trim={0cm 0 0 0},clip,width=0.8\textwidth]{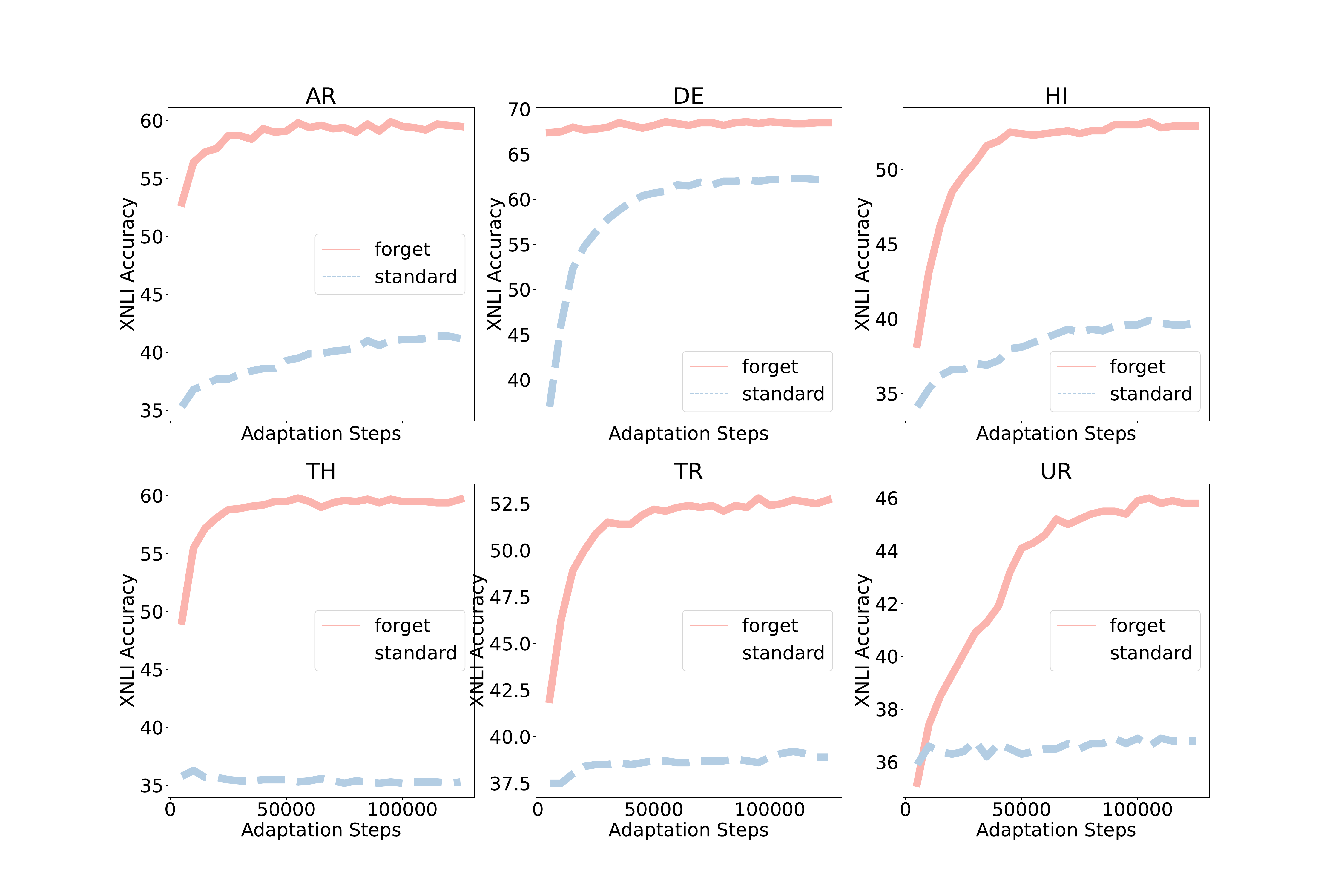}
    \caption{Adaptation curves on XNLI for individual languages: Arabic, German, Hindi, Thai, Turkish, and Urdu. Forgetting helps more languages that are distant to English (the pretraining language).}
    \label{fig:5M_125K}
\end{figure}

\begin{figure}
    \centering        
    \includegraphics[trim={0cm 0 0cm 4cm},clip,width=1\textwidth,left]{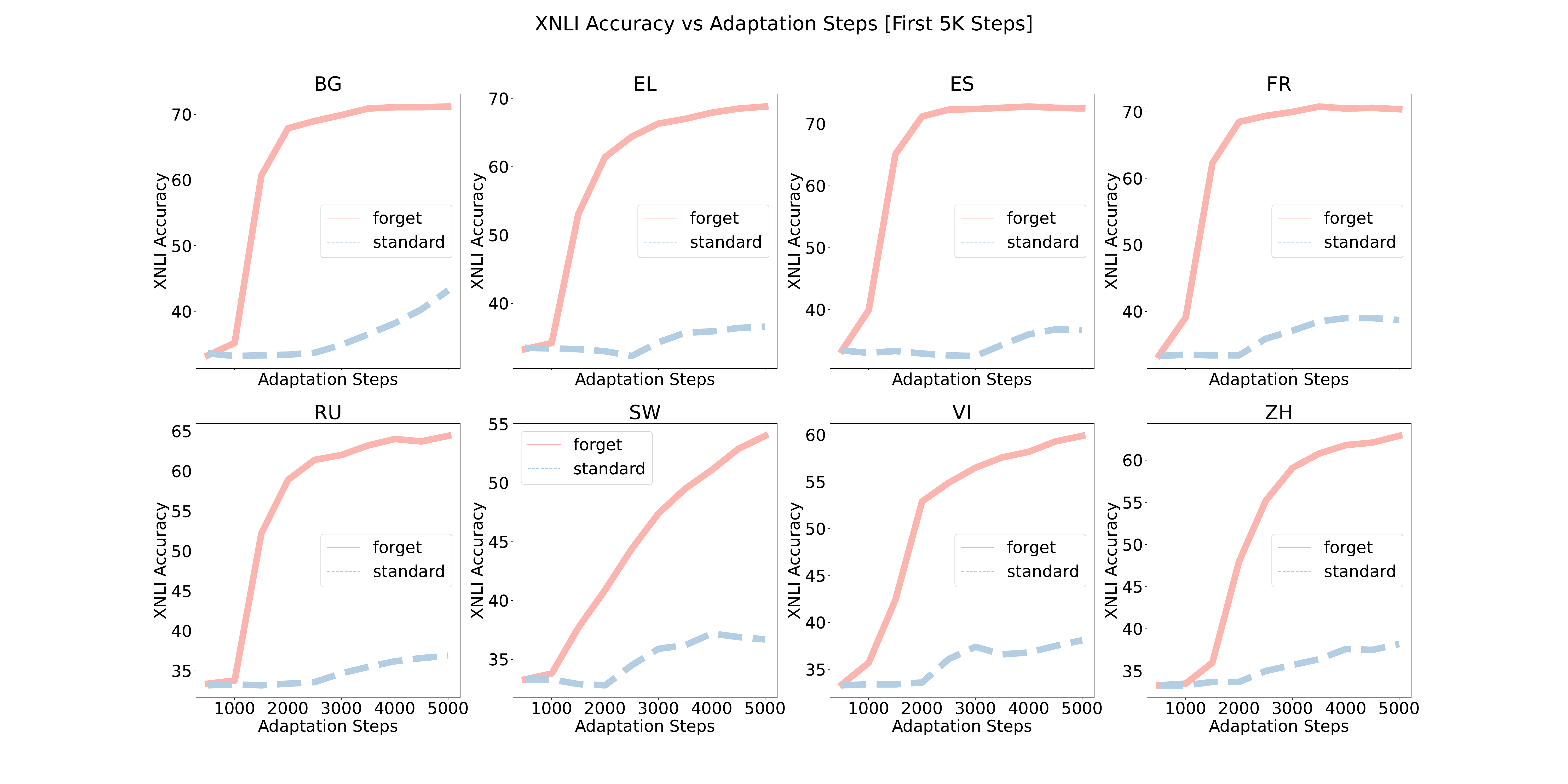}
    \caption{Adaptation curves on XNLI within 5K updates for individual languages: Bulgaria, Greek, Spanish, French, Russian, Swahili, Vietnamese and Chinese. Forgetting PLMs converge faster than standard PLMs.}
    \label{fig:5M_5K_more}
\end{figure}

\begin{figure}
    \centering        \includegraphics[trim={0cm 0 0 4cm},clip,width=1\textwidth,left]{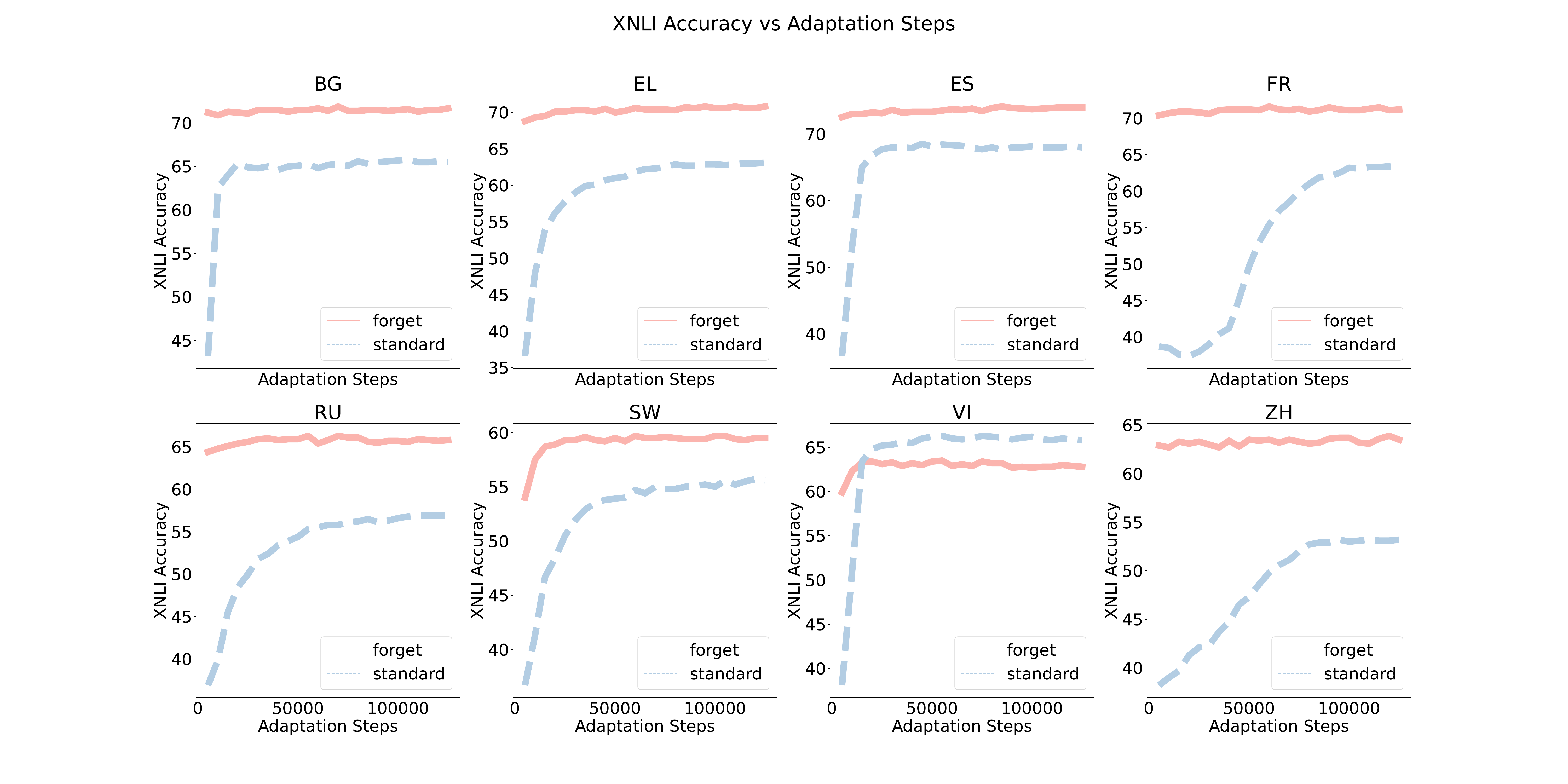}
    \caption{Adaptation curves on XNLI for individual languages: Bulgaria, Greek, Spanish, French, Russian, Swahili, Vietnamese and Chinese. Across all the languages except on Vietnamese, the forgetting PLMs reach a better performance level than their standard counterparts.}
    \label{fig:5M_125K_more}
\end{figure}

\Cref{fig:5M_125K} displays the adaptation curves for several languages (Arabic, German, Hindi, Thai, Turkish, and Urdu) during full training runs of 125,000 steps. This complements \Cref{fig:5M_5k}, which focuses on the first 5,000 steps. Similar convergence patterns can be observed for additional languages, as shown in ~\Cref{fig:5M_5K_more} and ~\Cref{fig:5M_125K_more}.

\section{More Detailed Analysis}
\subsection{Impact of Forgetting Frequency}
We would like to elaborate on our choice of forgetting frequency $K$. In our preliminary experiments, we tried $K=100, 1000, 5000$. We find $K=1000$ works well and thus sticks with it. Since we don't want to overtune the hyperparameters, we just use the same $K$ for all the experiments. We include the loss curves of $K=100$ and $K=5000$ here. We can see that both forgetting too frequently and forgetting too infrequently will hurt the performance. Too frequent forgetting leaves little time for the body to learn something meaningful (the pretraining loss stuck around $11$). Too sparse forgetting will make the body hard to adjust to the next forgetting, causing divergence as pretraining goes on.
\begin{figure}[htbp]
        \centering
        \includegraphics[scale=0.05]{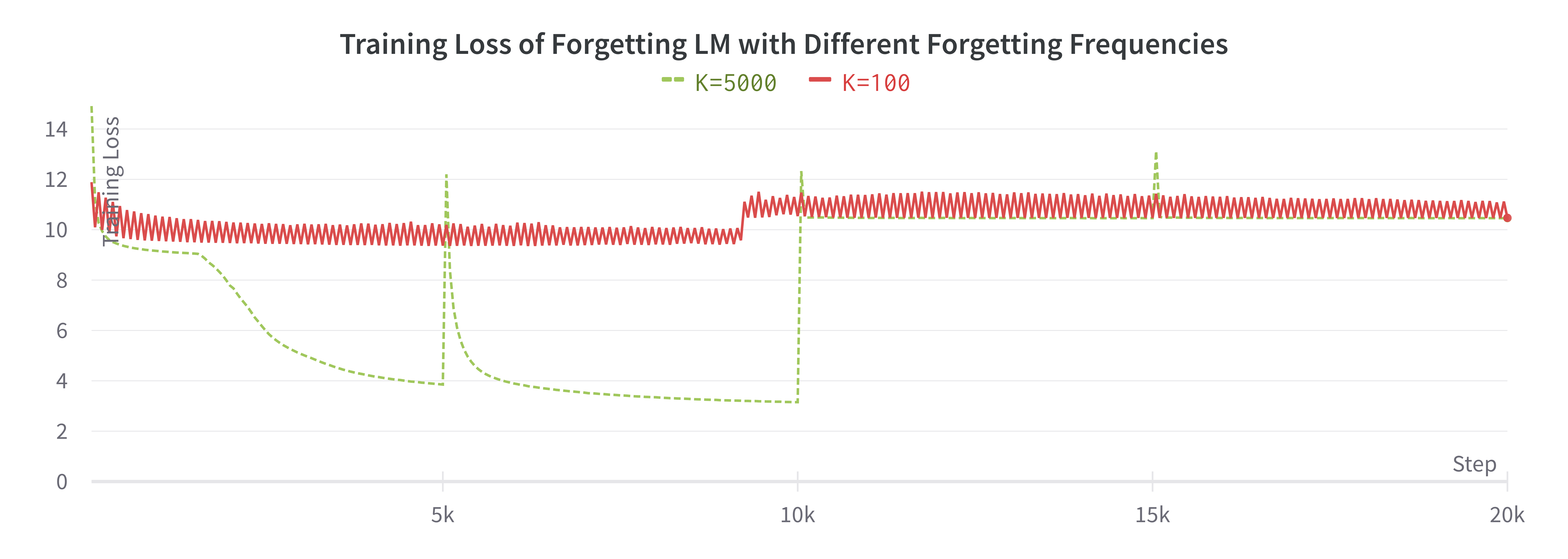}
    \caption{Impact of Forgetting Frequency.}
        \label{fig:freq}
\end{figure}

\subsection{Multilingual Pretraining and Forgetting}
Our work aim to have a flexible language model by pretraining with forgetting. No matter the pretraining corpus is monolingual or multilingual, this language model should easily generalise itself to unseen languages. This is different from the scenario of multilingual PLMs like XLM-R~\citep{conneau-etal-2020-unsupervised}, which requires seeing all the data for all languages from the scratch. Once done with pretraining and there is some new language distant from the pretraining languages to support, the multilingual PLMs might still struggle with zero-shot transfer as shown in several low-resources language research~\citep{ebrahimi2022americasnli,adelani2021masakhaner,adelani2022masakhaner}.

Nevertheless, we ran additional experiments on multilingual pretraining with forgetting. For a fair comparion, we trained a multilingual RoBERTa-base of the same model size as our monolingual model. %
Language Emb/Task Body Adaptation refers to separately adapting embeddings with 5M tokens of Thai unlabelled data and adapting body with English NLI data. Task Full Model Adaptation refers to adapting the full model with English NLI data. Note that Thai is already included in multilingual CC100 (6B tokens in the original dataset, 720M tokens in our subsampled dataset). We measure the zero-shot Thai XNLI Accuracy.

We can see that multilingual pretraining indeed helps cross-lingual transfer when the language is in the pretraining data. On the other hand, we can also observe that forgetting indeed lifts the adaptation performance:

\begin{itemize}
\item Comparing Row 3 and Row 4 (49.4 vs 55.0), we can see that, forgetting also helps adapt multilingual pretrained models.
\item Comparing Row 1 and Row 2 (35.3 vs 59.7), we can see that forgetting helps monolingual pretrained models a lot
\item XLM-R (base) outperform best our multilingual pretrained baselines (72.4 vs 60.0). This is no surprise due to its large pretraining corpus (10x our multilingual corpora) and model size (2x our multilingual model).
\end{itemize}

\begin{table}[]
\resizebox{\textwidth}{!}{
\begin{tabular}{l|l|l|l|l|l}
\toprule
\textbf{Pretrain Corpus} & \textbf{\#Langs} & \textbf{\#Params} & \textbf{Pretraining Method} & \textbf{Adaptation Framework} & \textbf{Acc} \\ 
\midrule
300GB English & 1 & 125M & Standard-RoBERTa (base) & Lang Emb/Task Body & 35.3 \\ 
300GB English & 1 & 125M & Forget-RoBERTa (base) & Lang Emb/Task Body & 59.7 \\ 
50GB Multilingual & 100 & 125M & Standard-RoBERTa (base) & Lang Emb/Task Body & 49.4 \\ 
50GB Multilingual & 100 & 125M & Forget-RoBERTa (base) & Lang Emb/Task Body & 55.0 \\ 
50GB Multilingual & 100 & 125M & Standard-RoBERTa (base) & Task Full Model & 60.0 \\ 
2.5TB Multilingual & 100 & 270M & XLM-RoBERTa (base) & Task Full Model & 72.4 \\ 
\bottomrule
\end{tabular}}
\end{table}

\subsection{Full Model Task Adaptation and Forgetting}
The language/task adaptation does not use any labelled data. It only uses the unlabelled data from the new language. In contrast, Standard adaptation relies on labelled data, which is expensive for a new downstream language.  In our case, we only found 
 Arabic NLI data. The amount of labelled data is not enough to adapt an English NLI model to Arabic NLI without proper regularization.

Our experimental setup follows~\citet{artetxe2020cross}, where standard-pretraining + language/task adaptation (MonoTrans) is shown to be competitive among a few baselines for zero-shot unsupervised cross-lingual transfer. On top of this finding, our proposed forgetting method can further improve the sample-efficiency of the language/task adaptation, reducing the amount of unsupervised data needed for the new language. This is motivated by a practical scenario where the new languages contain only several thousands of tokens to a few millions of tokens (e.g. the corpus for the new language might contain only 2-3 books).

\begin{table}[h]
\resizebox{\textwidth}{!}{
\begin{tabular}{l|l|l|l}
\toprule
\textbf{Method} & \textbf{Supervised Data} & \textbf{Unsupervised Data} & \textbf{Acc} \\ 
\midrule
standard pretraining + standard adapt & 6.7K & 0 & 32.8 \\ 
standard pretraining + language/task adaptation & 0 & 5M & 41.2 \\
forget pretraining + standard adaptation & 6.7K & 0 & 34.2 \\
forget pretraining + language/task adaptation & 0 & 5M & 59.7 \\
\bottomrule
\end{tabular}}
\end{table}

\subsection{Impact of Adaptation Data Amount}
Evidence of high sample efficiency can be found by comparing the performance drop of standard PLMs and forgetting PLMs when the adaptation data change from a high-data setting~\citep{artetxe2020cross,marchisio2022mini} to a low-data setting that our work is considering.

\begin{tabular}{lcc}
    \toprule
    \textbf{Method} & \textbf{Avg adaptation \#tokens} & \textbf{Avg XNLI performance} \\
    \midrule
    Standard (Kelly et al 2022 [27]) & 10.3B & 72 \\
    Standard (Artetxe et al 2022 [18]) & 569M & 66.7 \\
    Standard & 5M & 53.3 \\
    Forgetting & 5M & 62.7 \\
    \bottomrule
\end{tabular}

\end{document}